\documentclass[lettersize,journal]{IEEEtran}
\usepackage{amsmath,amsfonts}
\usepackage{array}
\usepackage[caption=false,font=normalsize,labelfont=sf,textfont=sf]{subfig}
\usepackage{textcomp}
\usepackage{stfloats}
\usepackage{url}
\usepackage{verbatim}
\usepackage{graphicx}
\usepackage{cite}
\usepackage[square,numbers,sort&compress]{natbib}

\usepackage{color}
\usepackage{mathptmx} 
\usepackage{times} 
\usepackage{amsmath} 
\usepackage{amssymb} 
\usepackage{graphicx}
\usepackage{stfloats}
\usepackage{bm}
\usepackage{balance}
\usepackage[linesnumbered,ruled,vlined,commentsnumbered]{algorithm2e}
\usepackage{booktabs}  
\usepackage{float}
\usepackage{url}
\DeclareMathAlphabet{\mathcal}{OMS}{cmsy}{b}{n}
\DeclareMathAlphabet{\mathcal}{OMS}{cmsy}{m}{n}
\usepackage{enumerate}
\usepackage[table]{xcolor}
\usepackage{etoolbox}

\usepackage{footnote}
\usepackage{amsthm}
\usepackage{multirow}
\usepackage[colorlinks=true,
            citecolor=blue,
            ]{hyperref}
\renewcommand\arraystretch{1}
\usepackage{caption}
\makeatletter
\def\ps@IEEEtitlepagestyle{%
  \def\@oddhead{%
    \hbox{%
      \hspace*{0.3\textwidth} 
      This article has been accepted by \href{https://doi.org/10.1002/rob.22505}{Journal of Field Robotics} 
      \hspace*{0.2\textwidth} 
      \footnotesize\thepage
    }%
  }%
  \def\@evenhead{\@oddhead}%
  \def\@oddfoot{}%
  \def\@evenfoot{}%
}
\makeatother

\begin{document}

\title{\LARGE \bf
ROLO-SLAM: Rotation-Optimized LiDAR-Only SLAM in Uneven Terrain with Ground Vehicle
}

\author{Yinchuan Wang$^{1}$, Bin Ren$^{1}$, Xiang Zhang$^{1}$, Pengyu Wang$^{2}$, Chaoqun Wang$^{1}$, Rui Song$^{1}$, Yibin Li$^{1}$, Max Q.-H. Meng$^{2}$        
\thanks{$^{1}$Yinchuan Wang, Bin Ren, Xiang Zhang, Pengyu Wang, Chaoqun Wang, Rui Song, Yibin are with School of Control Science and Engineering, Shandong University, Jinan, China.}%
\thanks{$^{2}$Pengyu Wang and Max Q.-H. Meng are with Shenzhen Key Laboratory of Robotics Perception and Intelligence and the Department of Electronic and Electrical Engineering, Southern University of Science and Technology, Shenzhen, China.}
}


\maketitle

\begin{abstract}
LiDAR-based SLAM is recognized as one effective method to offer localization guidance in rough environments. However, off-the-shelf LiDAR-based SLAM methods suffer from significant pose estimation drifts, particularly components relevant to the vertical direction, when passing to uneven terrains. This deficiency typically leads to a conspicuously distorted global map. In this article, a LiDAR-based SLAM method is presented to improve the accuracy of pose estimations for ground vehicles in rough terrains, which is termed Rotation-Optimized LiDAR-Only (ROLO) SLAM. The method exploits a forward location prediction to coarsely eliminate the location difference of consecutive scans, thereby enabling separate and accurate determination of the location and orientation at the front-end. Furthermore, we adopt a parallel-capable spatial voxelization for correspondence-matching. We develop a spherical alignment-guided rotation registration within each voxel to estimate the rotation of vehicle. By incorporating geometric alignment, we introduce the motion constraint into the optimization formulation to enhance the rapid and effective estimation of LiDAR's translation. Subsequently, we extract several keyframes to construct the submap and exploit an alignment from the current scan to the submap for precise pose estimation. Meanwhile, a global-scale factor graph is established to aid in the reduction of cumulative errors. In various scenes, diverse experiments have been conducted to evaluate our method. The results demonstrate that ROLO-SLAM excels in pose estimation of ground vehicles and outperforms existing state-of-the-art LiDAR SLAM frameworks.
\end{abstract}


\section{Introduction}
\label{sec: introduction}

\IEEEPARstart{L}{ocalization} is of utmost significance in the context of autonomous driving. 
It serves as the fundamental building block for safe and efficient navigation, enabling vehicles to precisely determine their position within their environment. 
For uneven terrain navigation, the movement of the vehicle is inevitably subject to fluctuations as it negotiates uneven terrain. The sensors rigidly mounted on the ground vehicle are particularly vulnerable to these movements, rendering off-terrain localization a particularly challenging endeavor.




Simultaneous localization and mapping (SLAM) technique allows one to localize ego-pose of sensors and meantime offers an environmental map. This approach provides an effective solution for localization in uneven and unknown environments. This study focuses on employing the LiDAR-based SLAM method for uneven terrain navigation. 
The LiDAR-based methods typically utilize consecutive LiDAR scans for point cloud registration on the local and global scale, allowing for the estimation of precise sensor ego-motion. These methods enjoy the benefits of insensitivity to environmental conditions, long sensing range and low measurement noise, particularly in outdoor uneven scenes. 


Off-the-shelf LiDAR-based SLAM incorporates typically two modules: the front-end LiDAR odometry and back-end mapping optimization \cite{latif2014robust}. The front-end provides initial pose estimation through frame-to-frame registration, while the back-end employs alignment and optimization methods on a global scale to refine the pose estimation and reconstruct the surroundings. This approach enables the framework to achieve coarse-to-fine localization. It is universal to adapt to normal scenes, such as urban flat roads. However, in uneven cases, the LiDAR-based SLAM method deployed on the ground vehicles suffers from non-negligible drifts of localization, leading to a distorted or oblique map.
The substantial reason is that the pose components in the vertical direction undergo significant changes due to the vehicle's body shaking in response to the terrain surface.
These changes directly lead to a reduction in the consensus set during frame-to-frame matching, resulting in incorrect initial pose estimation by the front-end. 
Although much efforts \cite{lego_loam,galeote2023gnd,chen2024ig} have been made and achieved notable improvements, this topic remains challenging and needs further enhanced solutions.
\begin{figure}[t]
\centerline{\includegraphics[scale=0.25]{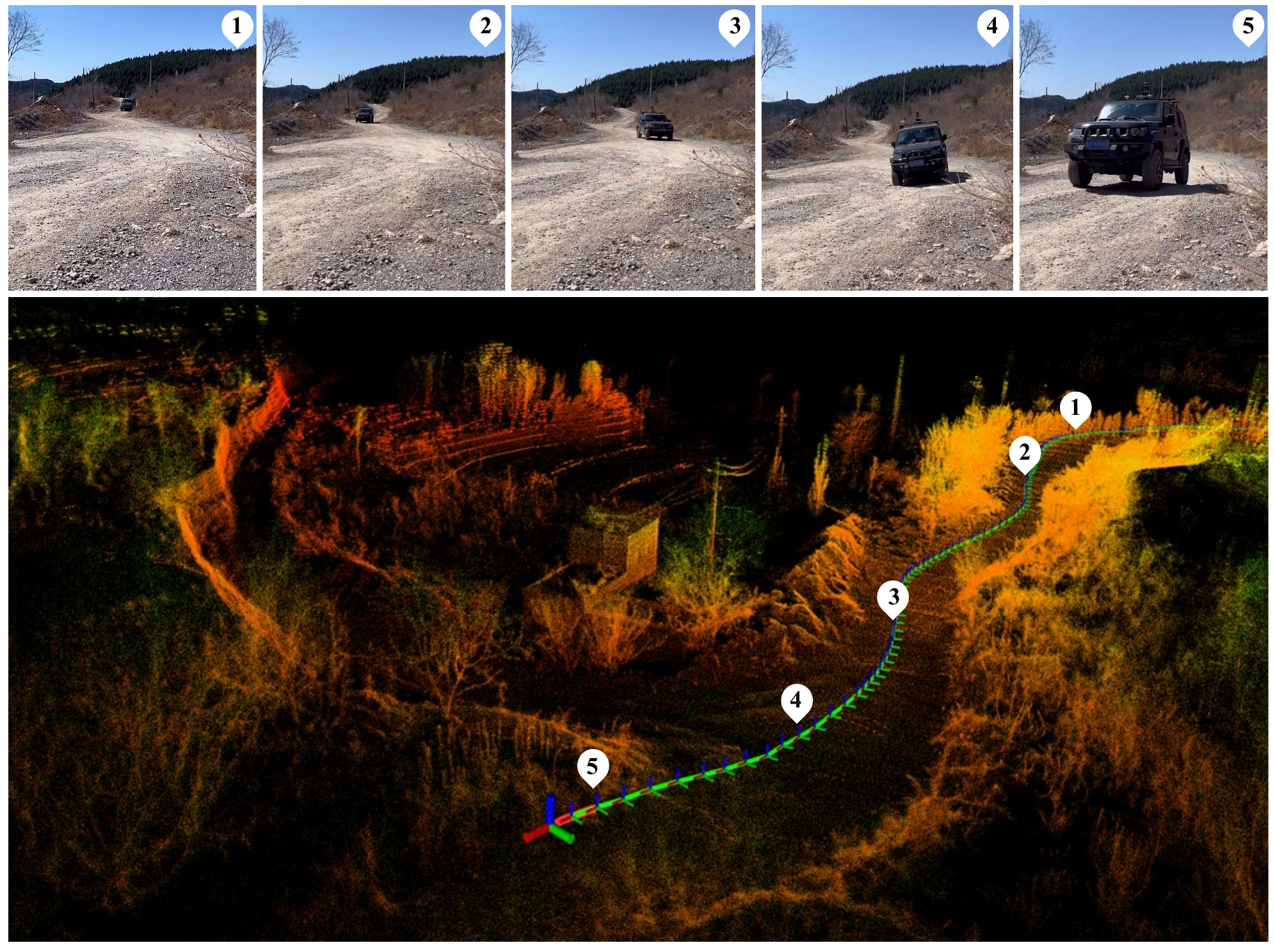}}
\caption{Top figures show a real vehicle moving in an off-road scenario. The bottom figure shows the point cloud map and the trajectory output by ROLO-SLAM.}
\label{fig: title}
\end{figure}
In response to this problem, we propose ROLO-SLAM: A \textbf{R}otation-\textbf{O}ptimized \textbf{L}iDAR-\textbf{O}nly SLAM framework aiming to reduce the pose drift in the vertical direction and estimate the ground vehicle's pose more precisely in uneven terrains. We divide the front-end into three separate modules, which is based on the observation of the vertical drift of pose estimation in rough terrain cases. In the front-end, the developed forward location prediction is used for coarse translation estimation to decouple the rotation and translation. 
Subsequently, the voxelization matching and rotation registration are utilized to independently estimate the precise rotation between two consecutive scans. 
A continuous-time-based translation estimation is leveraged to obtain more precise translation of scans. We then integrate our method into an efficient SLAM framework with the scan-to-submap alignment and global factor graph as back-end. 
Overall, our main contributions lie in the following aspects:

\begin{itemize}
    \item We leverage a forward location prediction to achieve a soft decouple between rotation estimation and translation estimation, which allows us to estimate the rotation and translation independently.
    
    \item In the front-end, we present a dual-phase paradigm for rotation and translation estimation using spherical alignment and continuous-time optimization. It aims to offer a precise initial pose for back-end optimization.
    
    
    \item A compact LiDAR SLAM framework is established by integrating scan-to-submap alignment and global factor graph optimization, which facilitates the localization of ground vehicles within uneven terrains.   
\end{itemize}

Extensive experiments are conducted to validate the efficacy of the proposed method and the results showcase that our method has the best unified performance compared with the state-of-the-art SLAM framework. Fig. \ref{fig: title} shows an example result and real-world snapshots in an off-road scene. 
In addition, the source code and video demonstration of our method are available\footnote{Code: \href{https://github.com/sdwyc/ROLO}{https://github.com/sdwyc/ROLO}\\ 
Video: \href{https://youtu.be/SeNuQmBAWFU}{https://youtu.be/SeNuQmBAWFU}}.

This article is organized as follows. Sec. \ref{related_work} discusses the related research. Sec. \ref{problem definition} addresses problem formulation and potential causes, followed by the SLAM system pipeline and detailed mechanism in Sec. \ref{ROLO system}. Subsequently, the conducted experimental evaluation is presented in Sec. \ref{experiments} and the corresponding results analysis is illustrated in Sec. \ref{result analysis}. Finally, Sec. \ref{conclusion} summarizes our conclusions and discusses future research directions.

\section{Related Work}
\label{related_work}
Although LiDAR-based SLAM with ground vehicles has spawned numerous practical applications, such as search and rescue, autonomous driving and subterranean exploration \cite{jian2022putn,ebadi2020lamp,xue2023traversability}, concerns persist regarding the vehicle's SLAM or localization in harsh environments. For the pioneering LiDAR-based method \cite{2014loam}, it becomes apparent that significant vertical drifts occur when the vehicle traverses uneven terrain, resulting in distorted and overlapping point cloud maps. To date, numerous studies have been conducted \cite{wang2021floam,lin2020livox_loam,chen2022direct} towards enhancing both accuracy and real-time capabilities of LiDAR-based SLAM. However, the reduction of vertical errors remains limited when these methods are implemented in those challenging environments.

One potential strategy to alleviate the aforementioned problem lies in promoting the accuracy of point registration. 
The Iterative closest point (ICP) algorithm \cite{pomerleau2013comparing} and its variant are generally used for achieving the alignment tasks in the LiDAR-based SLAM. For example, \cite{li2022gicp_loam} propose a refined LiDAR SLAM framework using generalized ICP (GICP) algorithm, which exploits the spatial voxelization to enhance the point-matching process. To improve the robustness towards LiDAR point cloud, \cite{chen2021ndtloam} leverages the normal distribution matching to replace the point-to-point matching by pursuing the maximum of joint probability. Recently, \cite{dellenbach2022cticp} present a CT-ICP method by adding a continuous time constraint to the optimization function to pursue smoother pose transformation. 
This approach demonstrates remarkable efficiency in dealing with variant terrain, making it a possible candidate for reducing vertical errors in harsh environments.

Furthermore, feature-based methods \cite{2023fastfeature,2023localinformation,2022linefeature} enhance the efficiency and efficacy of registration compared with the original ones by aligning representative features, such as edge features and planar features in the environments. 
\cite{wang2022fevo} develop FEVO-LOAM framework towards solving vertical drift problem, which enhances the feature extraction to capture the valid line, planar and ground feature points. 
Moreover, \cite{chen2022low} formulate a slope feature extraction method and incorporate it into a factor graph to achieve optimized pose estimation in uneven terrain. Nevertheless,  these methods all simultaneously estimate translation and rotation, leading to a vast solution space for optimization problems and hindering rapid convergence. 
\cite{lego_loam} propose a LeGO-LOAM method towards accurately estimating the pose in variant terrain. This method divides the estimation of the pose into several steps and optimizes the results using different features, which enables to shrinking of the solution space to guarantee the convergence quality of the solution. \cite{yang2020teaser} use the translation-invariant principle to achieve the rotation and translation registration independently. This method has been verified to decrease the optimization solution space and achieve improved results through numerous experiments.

Another approach to address the vertical drift problem is multi-sensor information fusion. 
This method typically leverages multi-scale features of environments to compensate for the absence of vertical observation using multiple sensors such as LiDARs, cameras, and IMUs \cite{chen2022eil,wang2022dlio}.
\cite{shan2020lio} propose a tightly-coupled LiDAR-inertial SLAM method, which uses the IMU and LiDAR to tightly track the pose transformation and adopts a factor graph to fuse these observations. 
SDV-LOAM \cite{2023sdv} leverages the abundant camera observation to achieve a semi-direct odometry in the front-end while the LiDAR sensor is used for executing sweep-to-map optimization in the back-end. 
However, of note is that in multi-sensor fusion, the integration of data from diverse sensors can introduce additional errors if precise calibration is not achieved. Therefore, careful calibration and synchronization of sensors are crucial to ensure accurate and reliable registration results.

With the soaring in popularity of deep learning (DL) techniques, researchers are increasingly acknowledging the potential of DL methods to tackle challenges in environmental sensing and data association.
The general methods \cite{pais20203dregnet,li2019net,chen2020overlapnet} leverage the deep neural network to accomplish complex matching and recognition tasks, which are under sparse sensor observation. 
For example, \cite{chen2019suma++} present SuMa++ to achieve high-quality mapping and odometry measurement tasks. SuMa++ extracts the semantic information by network and combines it with the surfel feature to construct a continuous semantic map. Similarly, \cite{deng2023nerf} and \cite{2023slamesh} exploit neural networks to render point clouds into the mesh map using state-of-the-art 3D reconstruction techniques. 
While DL-based approaches offer promising solutions for complex registration problems, balancing accuracy, efficiency, and robustness is essential for the widespread adoption of these techniques in real-world environmental sensing applications.



\begin{figure}[t]
\centerline{\includegraphics[scale=0.42]{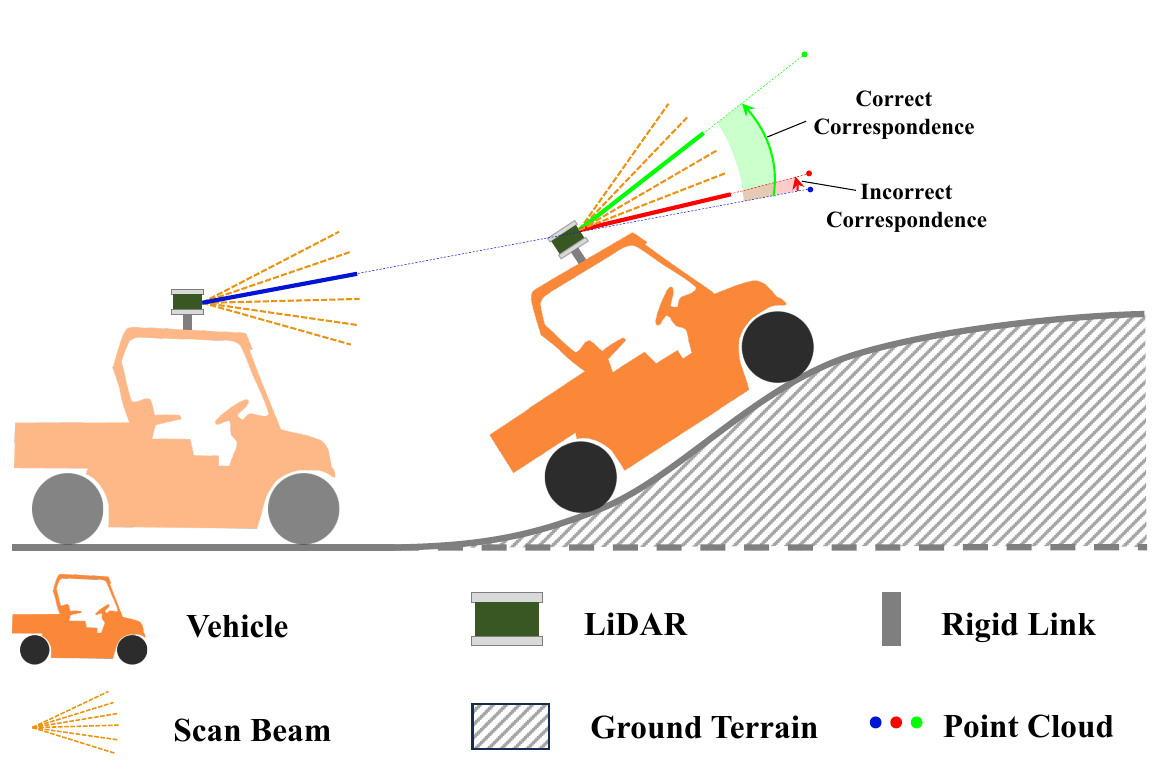}}
\caption{A simple case of suffering correspondence problems on uneven terrain.}
\label{fig: correspondence problem}
\end{figure}

\section{Problem Definition}
\label{problem definition}

\begin{figure*}[t]
	\centering
	\includegraphics[scale=0.45]{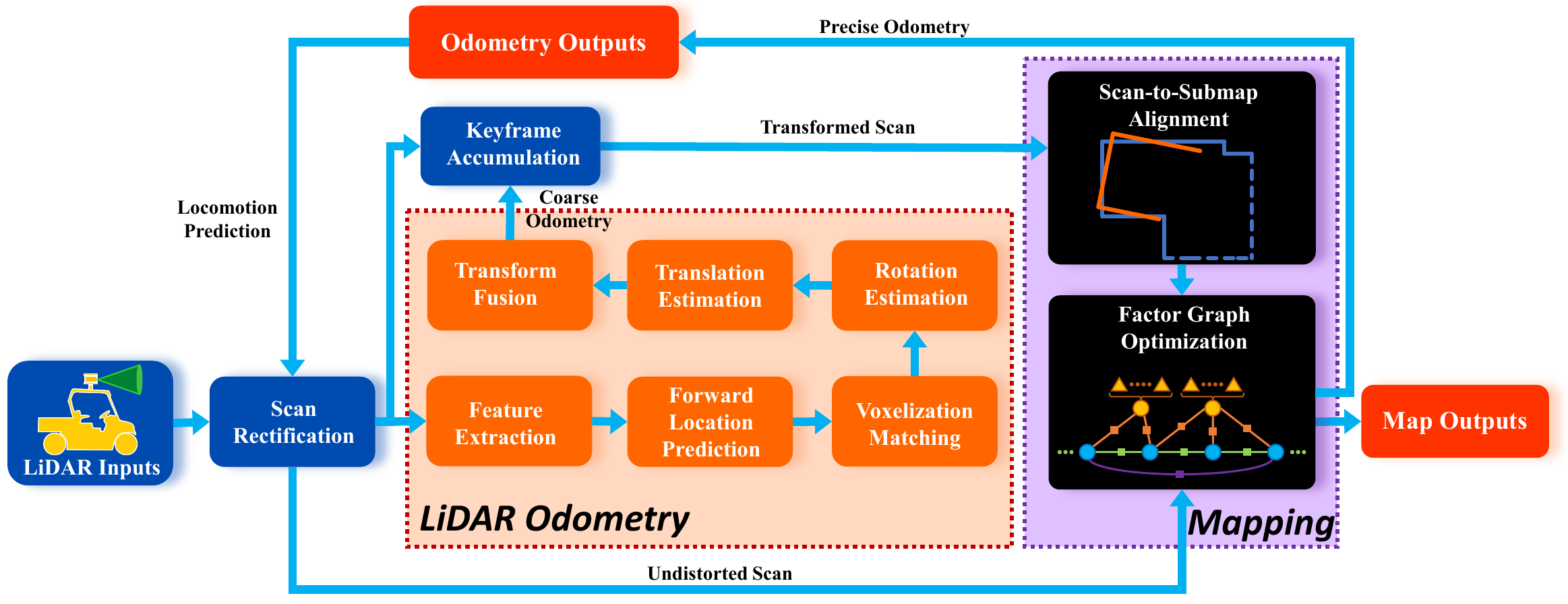}
	\caption{A system pipeline of our ROLO-SLAM, incorporating front-end LiDAR odometry module and the back-end mapping module.}
	\label{fig: system_overview}
\end{figure*}

In the world frame $\mathcal{W} \in \mathbb{R}^{3}$, 
we denote $\mathcal{B}$ and $\mathcal{L}$ as the vehicle and LiDAR frame respectively. The pose is represented by a transformation matrix $\mathbf{T} \in SE(3)$, abbreviated as $\left [ \mathbf{R}\  | \ \mathbf{t}  \right ]$.
Where $\mathbf{R} \in SO(3)$ represents the rotation matrix, $\mathbf{t} \in \mathbb{R}^{3}$ is translation vector.
Concerning a vehicle-LiDAR rigid connection system, the vehicle's world pose $^{\mathcal{B}}_{\mathcal{W}}\mathbf{T}$ can be derived from LiDAR's world pose $^{\mathcal{L}}_{\mathcal{W}}\mathbf{T}$, which is computed by:
\begin{equation}
\label{eq. connection}
^{\mathcal{B}}_{\mathcal{W}}\mathbf{T} = [{^{l_{2}}_{l_1}\mathbf{T}} \dots {^{l_{n}}_{l_n-1}\mathbf{T}}] ^{\mathcal{L}}_{\mathcal{W}}\mathbf{T},
\end{equation}
where $l_{i}, i\in\{1,2,\dots,n\}$ represent the links connecting vehicle and LiDAR. In addition, each scan $\mathcal{P}$ from LiDAR is composed of point set $\{\mathbf{p}_i \in \mathcal{P}\}$. $\mathcal{M}$ represents the point cloud map.

We assume that the rigid connection is always maintained between the vehicle and LiDAR.
Our objective is to determine LiDAR's pose $^{\mathcal{L}}_{\mathcal{W}}\mathbf{x}$ and then derive the vehicle's pose $^{\mathcal{B}}_{\mathcal{W}}\mathbf{x}$ using Eq. \ref{eq. connection}. 

Most LiDAR SLAM methods produce visible pose drifts in the vertical direction when deployed on the ground vehicle moving on uneven terrains. The main reasons for its inception lie in two aspects. 
In one respect, the presence of non-leveled terrain surface necessitates changes in the vehicle pose, especially in the roll and pitch direction. 
In such scenarios, LiDAR sensors exhibit large angular displacements in the vertical direction as the vehicle navigates uneven terrain. However, the vertical resolution of the LiDAR is limited, leading to the gradual accumulation of pose errors.

In another respect, most point cloud registration methods used in LiDAR SLAM exploit iterative optimization to approximate a solution.
Fig. \ref{fig: correspondence problem} shows a case of autonomous driving on uneven terrain. 
The light orange vehicle occupies the position at the last moment, while the dark orange one represents the current moment, denoted as $ \mathcal{B}_{t-1} $ and $ \mathcal{B}_{t} $, respectively. 
The point cloud produced by LiDAR $\mathcal{L}_{t-1}$ and $\mathcal{L}_{t}$ are $\mathcal{P}_{t-1}$ and $\mathcal{P}_{t}$, respectively. 
In such case, the pose transformation $\mathbf{T}_{t-1}^{t}$ from the last moment to the current can be calculated by
\begin{equation}
\label{eq: correspondence}
\mathbf{T}_{t-1}^{t} = \underset{\mathbf{T} }{\mathrm{argmin}} \sum \mathbf{p}_{t}^{i} - \mathbf{T} \mathbf{p}^{j}_{t-1},\ \mathbf{p}_{t}^{i} \in \mathcal{P}_t, \mathbf{p}^{j}_{t-1} \in \mathcal{P}_{t-1},
\end{equation}
where $\left \langle \mathbf{p}^{i}_{t}, \mathbf{p}^{j}_{t-1}\right \rangle$ is called a correspondence, whose correctness has direct influence with the solution quality of Eq. \ref{eq: correspondence}. The correct correspondence indicates that the point transformation $\mathbf{T}$ is consistent with pose transformation $\mathbf{T}_{t-1}^{t}$ while the incorrect correspondence indicates that they are inconsistent.
However, with the intense shaking from the vehicle in uneven terrain cases, the points from two scans are prone to produce incorrect correspondence. As shown in Fig. \ref{fig: correspondence problem}, the blue point $B$ is from $\mathcal{P}_{t-1}$ while the green $G$ and red $R$ ones are from $\mathcal{P}_{t}$. Concerning the conventional closest match rule, the blue point is matched with the red point to generate a correspondence. The green point is farther from the blue point, hence they are not matched. But the correspondence $\left \langle B, R \right \rangle$ is incorrect since it is unable to reflect the pitch angle change between $\mathcal{B}_{t-1} $ and $ \mathcal{B}_{t}$. $\left \langle B, G \right \rangle$ is the correct correspondence that effectively reflects the change.
Furthermore, the optimization variable $\mathbf{T}$ needs to account for changes in 6D pose instead of the way only considering 3D poses in flat scenes. 
These factors lead to a reduction in the size of the maximum consensus set, thereby resulting in inferior solutions. 

In this study, in the case of vehicle localization in uneven terrain, our target problem focuses on how to improve vehicle localization accuracy by using a single LiDAR sensor. To this end, we reconstruct the entire front-end and divide it into several modules to refine the pose estimation. Subsequently, we introduce scan-to-submap alignment and factor graph to optimize the vehicle's pose within the map.

\section{Methodology}\label{ROLO system}

\subsection{System Pipeline of ROLO-SLAM}
The architecture of ROLO-SLAM is illustrated in Fig. \ref{fig: system_overview}. The developed framework is composed of two components: the front-end LiDAR odometry module and the back-end mapping module. 
Initially, the LiDAR scan data is rectified to correct motion distortion utilizing odometry data sourced from the back-end. 
In the front-end, geometric features are extracted based on their edge and planar characteristics that are identified through the metric of smoothness \cite{2014loam}. Subsequently, the forward location prediction is developed for rapid preliminary estimation of the LiDAR translation, which facilitates the loose decoupling of rotation and translation. 
This process is elaborated in Sec. \ref{state propagation}. 
Correspondences are determined using the developed voxelization approach. The rotation and translation are independently determined, where the rotation is registered by spherical alignment model while the translation is obtained based on continuous time optimization, detailed in Sec. \ref{rotation estimation} and Sec. \ref{translation estimation}. 
Furthermore, the back-end initiates by aggregating keyframes to construct submaps. These are leveraged to a scan-to-submap alignment, and further optimize the LiDAR's global pose and point cloud map using a factor graph, as detailed in Sec. \ref{mapping and lc}.


\subsection{Forward Location Prediction}
\label{state propagation}
\begin{figure}[t]
\centerline{\includegraphics[scale=0.5]{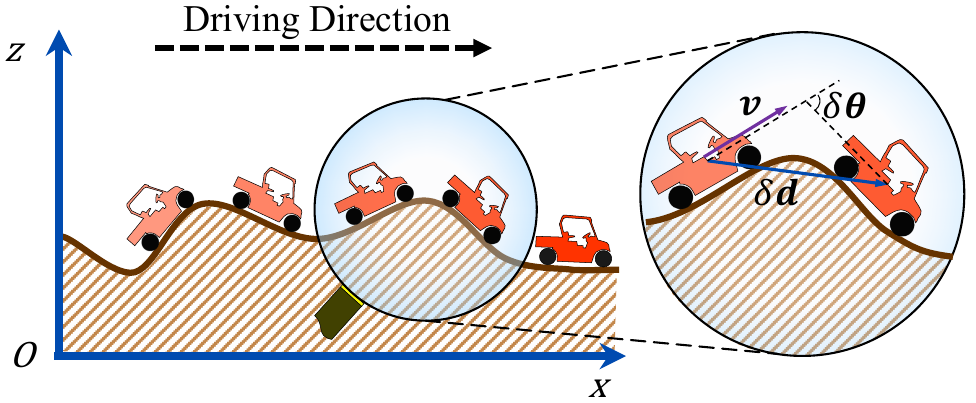}}
\caption{Snapshots of vehicle driving in uneven terrain. Here, the poses of any two vehicles are recorded at the same time intervals.}
\label{fig: state_propagation}
\end{figure}

In the front-end, we decouple the estimation of the rotation and translation from consecutive scans. 
It is achieved by eliminating the translation differences through a forward location prediction.
Fig. \ref{fig: state_propagation} displays snapshots of the vehicle during the same scan interval on the \textit{xoz} plane.
Given the $k$-th LiDAR scan and the corresponding robot velocity at that scan  $\mathbf{v}_k$, the robot velocity corresponding to the previous scan is denoted 
 as $\mathbf{v}_{k-1}$. As the time interval is sufficiently small, it can be expected a consistent linear velocity between two successive scans with off-the-shelf velocity control technology, i.e., $\mathbf{v}_k \sim \mathbf{v}_{k-1}$. The translation distance between two scans can be deemed the same, i.e., $\delta \mathbf{d}_{k-1} \sim \delta \mathbf{d}_{k}$. 
Denote the vehicle's \textit{k}-th location as the $\mathbf{t}_k$ after undergoing \textit{k} LiDAR scan $\mathcal{P}_{0:k}$. Upon receiving $\mathcal{P}_{k+1}$, we can estimate the vehicle location $\mathbf{t}_{k+1}$ by
\begin{equation}
\label{eq: state}
\mathbf{t}_{k+1} = \mathbf{t}_{k} + \frac{\tau_{k+1}-\tau_{k}}{\tau_{k}-\tau_{k-1}}  (\mathbf{t}_{k}-\mathbf{t}_{k-1}),
\end{equation}
where the $\tau_{i}$ represents the timestamp of $i$-th scan. 
When the scan $\mathcal{P}_{k+1}$ arrives, the vehicle's position at time $\tau_{k+1}$ is pre-estimated using Eq. \ref{eq: state} to form a coarse constraint to bound the translation differences between the scans $\mathcal{P}_{k}$ and $\mathcal{P}_{k+1}$. 
This procedure paves the way for independent subsequent estimations of rotation and translation for $\mathcal{P}_{k}$ and $\mathcal{P}_{k+1}$.




For the rotation of the vehicle, the pitch angle of the vehicle $\theta_i$ is constrained by the ground surface, which is more prone to be influenced by the undulation of the ground surface compared with the translation. 
In real-world applications, the undulation of ground is typically unknown and nonlinear; therefore, the pitch angle change $\delta \theta$ is hard to hold the same value facing uneven terrains during vehicle moving. The same thing is also applied in the analysis of the roll angle. We hereby do not estimate the rotation as the way in translation estimation. 

Traditional registration methods often intertwine rotation and translation estimations, thereby obscuring the distinct challenges associated with each, leading to potential inaccuracies in vehicle attitude and position.
By introducing forward location prediction, we establish a coarse estimation of the translation between consecutive LiDAR scans. This decouples the translation and rotation estimation in the front-end, which builds a consistent baseline of scan locations and is expected to improve the fidelity of vehicle rotation estimation.

\subsection{Voxelization Matching and Rotation Registration}
\label{rotation estimation}

It is challenging to accurately identify the point-to-point correspondences between scans. To cope with this problem, we first propose to use Gaussian voxel map. 
With a little abuse of notation, we represent the two adjacent scans $\mathcal{P}_{i-1}$ and $\mathcal{P}_{i}$ by the source point cloud $\mathcal{P}_{s}$ and the target point cloud $\mathcal{P}_{t}$, respectively.
The Gaussian voxel map is constructed as described in Alg. \ref{alg: voxelization} in the coordinate system of $\mathcal{P}_t$. We establish an empty voxel map $\mathcal{V}$ and a voxel index set $\mathcal{I}(\mathcal{V})$ to store the index value of voxel $\mathbf{m}_k \in \mathcal{V}$. Each point $\mathbf{p}_t \in \mathcal{P}_t$ is assigned a specific voxel, whose index is calculated by
\begin{align}
\mathbf{p}^{+}&= \frac{\mathbf{p}_t- \mathbf{p}_{min}}{{R}_{es}}, \\
\label{eq: voxel index}
voxel\_index &= \left [ 1,  W_{\mathcal{V}}, W_{\mathcal{V}} \cdot H_{\mathcal{V}} \right ]^{\top} \cdot  \mathbf{round}(\mathbf{p}^{+}),
\end{align}
where $\mathbf{p}_{min}$ is a reference point whose coordinate values in different directions $[x_{min},y_{min},z_{min}]$ are separately designated as the lowest coordinate values of the points in  $\mathcal{P}_t$.
$W_{\mathcal{V}}$, $H_{\mathcal{V}}$ and ${R}_{es}$ respectively represent the voxel map length, height, and resolution. $\mathbf{round}(\cdot)$ represents the rounding operation.
We represent any spatial point $\mathbf{p}_i \in \mathbb{R}^{3}$ suffered a white Gaussian noises as
\begin{algorithm}[t]
\caption{Voxelization}
\label{alg: voxelization}
\SetKwProg{Fn}{Function}{:}{}
\Fn{$\mathbf{Voxelization}(\mathcal{P}_{t})$}{
$\mathcal{V} \gets \emptyset$ \;
\For{$\mathbf{p}_{i} \in \mathcal{P}_{t}$}{
    $voxel\_index \gets \mathbf{CalculateIndex}(\mathbf{p}_i, \mathcal{V})$\;
    \If{$voxel\_index \notin \mathcal{I}(\mathcal{V})$}{
        $\mathcal{I}(\mathcal{V}) \gets \mathcal{I}(\mathcal{V}) \cup voxel\_index$\;
    }
    $\mathcal{V}[voxel\_index].N \gets \mathcal{V}[voxel\_index].N+1$\;
    $\mathcal{V}[voxel\_index].\bar{\mathbf{p}} \gets \mathcal{V}[voxel\_index].\bar{\mathbf{p}}+\mathbf{p}_{i}$\;
    $\mathcal{V}[voxel\_index].\bar{\bm{\Omega}} \gets \mathcal{V}[voxel\_index].\bm{\Omega}+\bm{\Omega}_i$\;
}
$[\mathcal{V}.\bar{\mathbf{p}}, \mathcal{V}.\bm{\Omega}] /= \mathcal{V}.N$\;
\Return{$\mathcal{V}$}
}
\end{algorithm}
\begin{equation}
\mathbf{p} _i \sim \mathcal{N} (\hat{\mathbf{p} _i}, \bm{\Omega} _i),
\end{equation}
where $\hat{\mathbf{p}_i}$ is location of the point. $\bm{\Omega}_i$ is the covariance matrix of the Gaussian white noise. 
For the target points, each voxel not only encapsulates a cluster of spatial points but also embodies these points through a Gaussian distribution. 
For each voxel $\mathbf{m}_k$, a Gaussian distribution is leveraged to approximate the spatial feature of points in there. This Gaussian distribution can be described as
\begin{align}
    m_k \sim \mathcal{N} &(\bar{\mathbf{p} } _k, \hat{\bm{\Omega} } _k ),\\
    \bar{\mathbf{p}}_k = \frac{\underset{i}{\sum} {\mathbf{p}}_i}{N_k}, & \    \bar{\bm{\Omega} } _k = \frac{\underset{i}{\sum} \bm{\Omega}_i}{N_k},
\end{align}
where $\bar{\mathbf{p}}_k$ and $\bar{\bm{\Omega} }_k$ represent the mean location of the points in $\mathbf{m}_k$ and covariance matrix, respectively. $N_{k}$ is the number of points in $\mathbf{m}_{k}$. This process is elaborated in Lines 8-12, Alg. \ref{alg: voxelization}.

Based on the voxelization, we hereby avoid directly considering the point-to-point correspondence. Instead, we seek for the pairs $\left \langle \mathbf{p}_{s}^{i}, \mathbf{m}_{k}\right \rangle$ consisting of a source point $\mathbf{p}_{s}^{i} \in \mathcal{P}_s$ and a target voxel $\mathbf{m}_{k} \in \mathcal{V}$ within this study. Based on the forward location prediction, we get the translation information. Then, we align the sensor centers of two consecutive scans at the same origin and start the matching, which is delineated in Alg. \ref{alg: matching}. We first establish the voxel map $\mathcal{V}$ following Alg. \ref{alg: voxelization} for $\mathcal{P}_{t}$. Then, the index of each source point $\mathbf{p}_s^i \in \mathcal{P}_{s}$ corresponding to $\mathcal{V}$ is calculated by Eq. \ref{eq: voxel index}, as indicated in Line 4, Alg. \ref{alg: matching}.
As illustrated in Alg. \ref{alg: matching} Lines 5-7, the matching rule is that a correspondence pair $\left \langle \mathbf{p}_{s}^{i}, \mathcal{V}[voxel\_index]\right \rangle$ is formed when $voxel\_index$ of $\mathbf{p}_{s}^i$ exists in $\mathcal{I}(\mathcal{V})$ of $\mathcal{V}$. Note that the voxel in $\mathcal{V}$ containing insufficient points is not eligible for matching with any source points. $N^{+}$ is a pre-setting threshold of the least point number. 
This ensures that only voxels that adequately represent the local geometry are considered for establishing correspondences.

\begin{algorithm}[t]
\caption{Matching}
\label{alg: matching}
\KwIn{Source point cloud: $\mathcal{P} _s$, Target points cloud: $\mathcal{P}_t$, Point corresponding $\mathcal{C}$}
\SetKwProg{Fn}{Function}{:}{}

$\mathcal{C} \gets \emptyset$ \;
$\mathcal{V} \gets \mathbf{Voxelization}(\mathcal{P}_t)$ \;
\For{$ \mathbf{p}^i_s \in \mathcal{P}_{s}$}{
    $voxel\_index \gets \mathbf{CalculateIndex}(\mathbf{p}^{i}_s, \mathcal{V})$\;
    \If{$voxel\_index \in \mathcal{I}(\mathcal{V})$ \textbf{and} $\mathcal{V}.N \ge N^+$}{
        $\mathcal{C} \gets \mathcal{C} \cup \textbf{make\_pair}(\mathbf{p}^i_s, \mathcal{V}[voxel\_index])$\;
    }
}
\Return{$\mathcal{C}$}
\end{algorithm}

\begin{figure}[t]
\centerline{\includegraphics[scale=0.4]{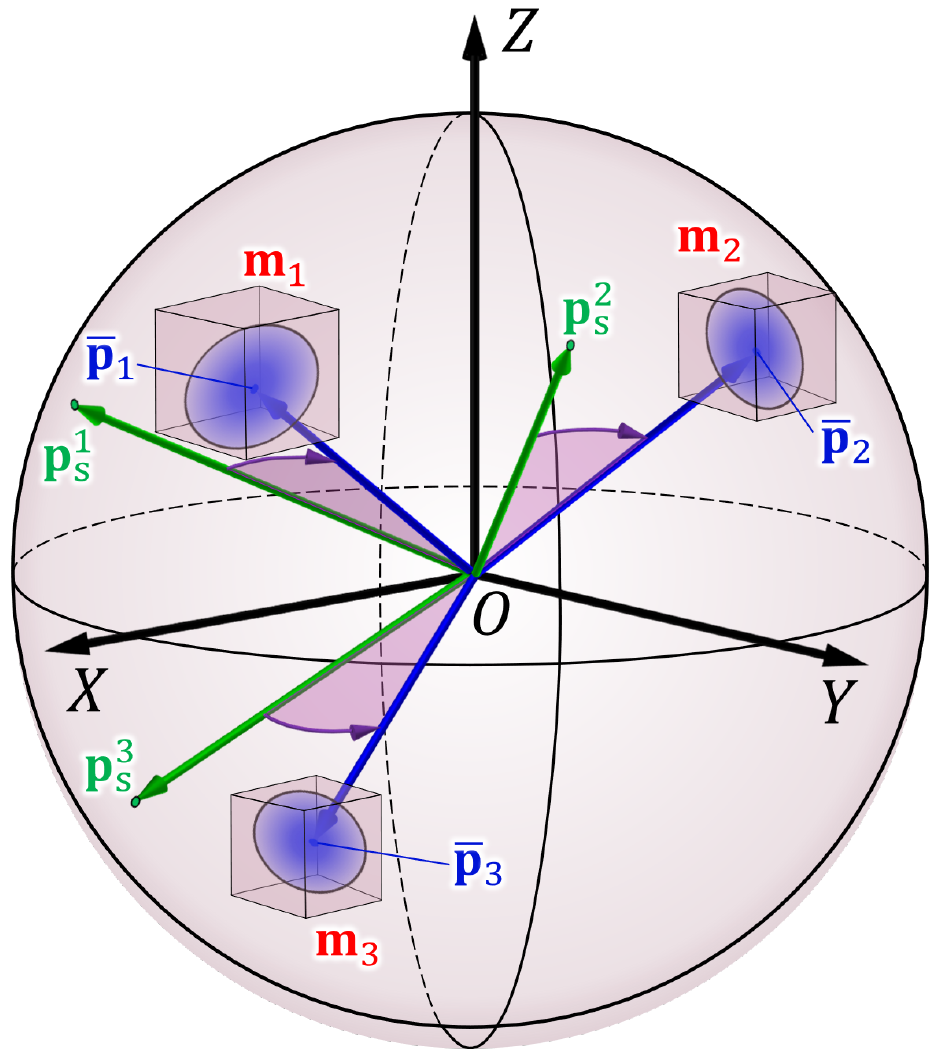}}
\caption{Rotation alignment model. The green points are source points in $\mathcal{P}_s$ while the blue ellipse spheres are the Gaussian distribution in $\mathbf{m}_k$. The purple arrows represent the possible rotated direction.}
\label{fig: muti_sphere}
\end{figure}

Now we obtained the correspondence $\mathcal{C}$ between the source points and the target voxels. Then, the rotation between $\mathcal{P}_s$ and $\mathcal{P}_t$ is obtained by attempting to align the points of $\mathcal{P}_s$ with the mean location $\bar{\mathbf{p}}_k$ of the associated Gaussian distribution stored in a voxel. The rotation alignment model is demonstrated by Fig. \ref{fig: muti_sphere}. Through forward location prediction, the sensor centers of two consecutive scans are positioned at the same origin $O$. Rotating a point cloud can be conceptualized as each point sliding along a spherical surface, with the LiDAR at the center and the distance to the point as the radius. 
Various source points $\mathbf{p}_{s}^i \in \mathcal{P}_{s}$ slide along the sphere's surface to align with the mean location $\bar{\mathbf{p}}_{k}$ of the Gaussian distribution, which is stored in the voxel $\mathbf{m}_{k}$. This alignment process estimates the sensor rotation between $\mathcal{P}_{s}$ and $\mathcal{P}_{t}$. To this end, the whole rotation alignment is expressed by 
\begin{equation}
\label{eq: rotation reg}
\mathbf{R}  = \underset{\mathbf{R} }{\mathrm{argmin} }\ \sum \sphericalangle   (\bar{\mathbf{p}}_k, \mathbf{R} \mathbf{p}^{i}_s),
\end{equation}
where $\sphericalangle(\cdot)$ represents the spherical angle between the associated point $\mathbf{p}^{i}_s$ and the mean $\bar{\mathbf{p}}_k$ of corresponding Gaussian distribution, with respect to the origin $O$. $\sphericalangle(\cdot)$ represents an angular metric, whereas the coordinates of spatial points are represented by distance metrics in this study. Subsequently, we propose a transformation that expresses $\sphericalangle(\cdot)$ in terms of a distance metric and derives the optimal rotation estimation via an optimization objective function formulated using the Mahalanobis distance.

\begin{figure}[t]
\centerline{\includegraphics[scale=0.17]{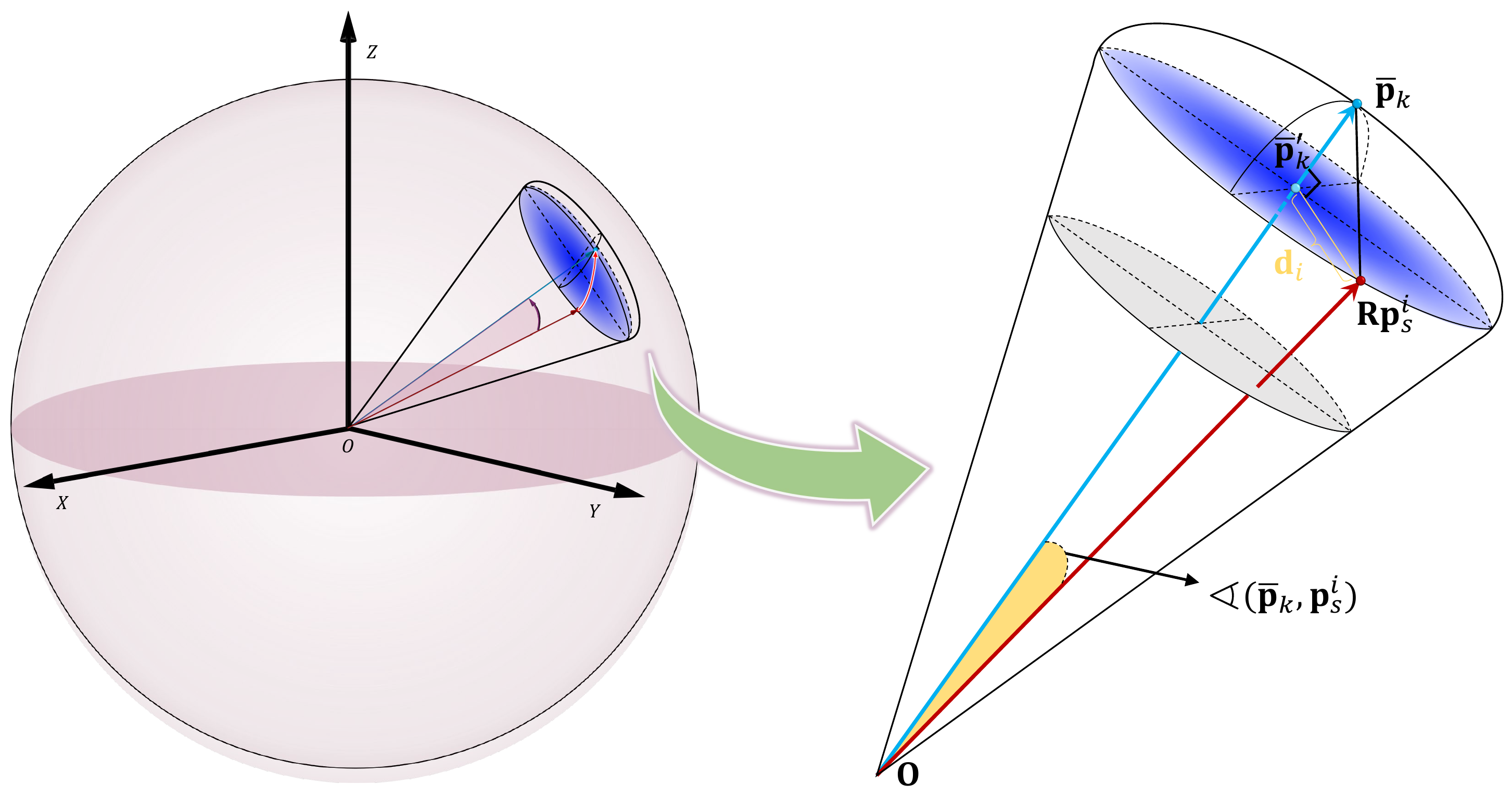}}
\caption{A demonstrated example for rotation alignment. The blue plane represents the projection of the Gaussian distribution with mean location $\bar{\mathbf{p}}_k$.}
\label{fig: rotation registration}
\end{figure}
\begin{figure}[t]
\centerline{\includegraphics[scale=0.5]{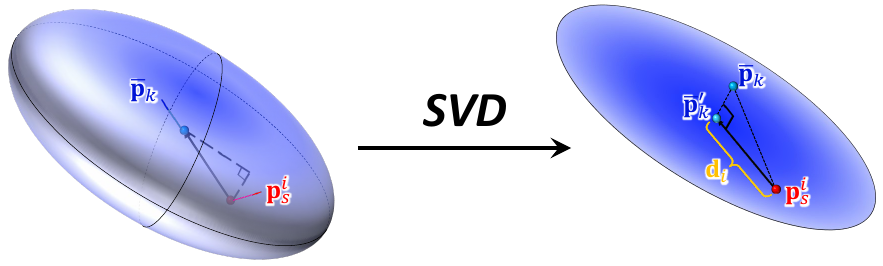}}
\caption{Transformation of the spatial manifold representing the 3D Gaussian distribution by employing an SVD regularization.}
\label{fig: svd}
\end{figure}

To conceptually illustrate the transformation,  Fig. \ref{fig: rotation registration} illustrates an example of aligning the source point $\mathbf{p}^{i}_{s}$ with the mean $\bar{\mathbf{p}}_{k}$ of the associated Gaussian distribution. 
Here, $\mathbf{R}\mathbf{p}^{i}_s$ represents the rotated source point by $\mathbf{R}$, and $\bar{\mathbf{p}}^{\prime}_{k}$ represents a projection of $\bar{\mathbf{p}}_{k}$ onto the spherical tangent plane that intersects $\mathbf{R}\mathbf{p}^{i}_s$. We have
\begin{align}
\sphericalangle   (\bar{\mathbf{p}} _k, \mathbf{R} \mathbf{p}^{i}_s) & = \arcsin \frac{\left \| \mathbf{d}_i \right \| }{\left \| \bar{\mathbf{p} }^{\prime}_k \right \| }, \\
\label{eq: metric trans}
\sphericalangle   (\bar{\mathbf{p}} _k, \mathbf{R} \mathbf{p}^{i}_s) & \propto  \left \| \mathbf{d}_i \right \| ,  s.t., \sphericalangle   (\bar{\mathbf{p}}^{\prime}_k, \mathbf{R} \mathbf{p}^{i}_s) \le \frac{\pi}{2},
\end{align}
where $\mathbf{d}_i = \bar{\mathbf{p}}^{\prime}_k - \mathbf{R} \mathbf{p}^{i}_s$. Eq. \ref{eq: metric trans} reveals the transformation from angle metric to distance metric. $\mathbf{p}^{\prime}_k$ can be calculated by
\begin{equation}
\bar{\mathbf{p}}^{\prime}_k = (\left \| \bar{\mathbf{p}} _k \right \|- \frac{(\bar{\mathbf{p}}_k - \mathbf{R} \mathbf{p}^{i}_s)\cdot  \bar{\mathbf{p}}_k}{\left \| \bar{\mathbf{p}}_k \right \| }) \mathbf{n}_k,
\end{equation}
where $\mathbf{n}_k$ represents the unit vector of $\bar{\mathbf{p}}_k$. Of note is that $\mathbf{d}_i$ is essentially influenced by a Gaussian white noise. We construct the covariance of this Gaussian noise as
\begin{equation}
\bm{\Omega}^{\prime}_\mathbf{R} = \bm{\Omega}_k + \mathbf{R} ^{\top}\bm{\Omega}^{i}_s\mathbf{R},
\end{equation}
where $\bm{\Omega}_k$ and $\bm{\Omega}^{i}_s$ are noise covariance matrix of $\bar{\mathbf{p}}_k$ and $\mathbf{p}^{i}_s$, respectively. We then perform singular value decomposition (SVD) on $\bm{\Omega}^{\prime}_\mathbf{R}$ and the covariance $\bm{\Omega}_\mathbf{R}$ is reconstructed as
\begin{equation}
\bm{\Omega}_\mathbf{R} = \mathbf{U}  \begin{bmatrix}
 \lambda_{max} &  & \\
  & \lambda_{max} & \\
  &  & \backepsilon   
\end{bmatrix} \mathbf{V}.
\end{equation}
$\lambda_{max}$ is the maximum of eigen values while $\mathbf{U}$ and $\mathbf{V}$ are obtained by SVD operation. Given that $\left \| \mathbf{d}_i \right \|$ quantifies the radial distance between $\mathbf{R}\mathbf{p}^{i}_{s}$ and $\mathbf{\bar{p}}^{\prime}_{k}$ on the spherical plane, our primary focus is on the discrepancies in radial distances. Conversely, the axial distance between $\mathbf{\bar{p}}^{\prime}_{k}$ and $\mathbf{\bar{p}}_{k}$ is less relevant to rotational registration. Therefore, SVD is employed to regularize the covariance, eliminating the influence of axial distances. This regularization can be interpreted as a dimensionality reduction of the Gaussian process, whereby the data is mapped from a 3D ellipsoidal manifold to a 2D elliptical manifold as shown in Fig. \ref{fig: svd}.

Thus, the rotation $\mathbf{R}$ can be calculated by
\begin{equation}
\label{eq: rot_opt}
\mathbf{R}=\underset{\mathbf{R} }{\mathrm{argmin} }\ {\sum \left \| \mathbf{d} _i \right \|}_{\bm{\Omega} ^{-1}_\mathbf{R}},
\end{equation}
where ${\left \| \cdot \right \|}_{\bm{\Omega}^{-1}_\mathbf{R}}$ represents the Mahalanobis distance while $\bm{\Omega}_\mathbf{R}$ is the covariance matrix. Eq. \ref{eq: rot_opt} can be solved iteratively using optimization algorithms, including Gaussian-Newton (GN) and Levenberg–Marquardt (LM).

\subsection{Continuous-Time-Based Translation Optimization}
\label{translation estimation}

Up to this point, we have obtained the rotation matrix $\mathbf{R}$ and a coarse translation estimation $\mathbf{t}$. To further optimize the translation, we design an objective function that incorporates a continuous-time-based translation constraint. 
This constraint is derived from the continuous uniform motion model of the vehicle between two successive LiDAR scans.
To make the presentation more concise, we introduce two symbols: $\breve{\mathbf{T}}$ and $\bar{\mathbf{T}}$, denoting the simple transformation matrixes that singly consider rotations and translations, i.e., $\breve{\mathbf{T}}=\left [ \mathbf{I}\  | \ \mathbf{t}  \right ]$ and $\bar{\mathbf{T}}=\left [ \mathbf{R}\  | \ \mathbf{0}  \right ]$.
Then, the translation optimization equation containing the objective function is expressed as follows:


\begin{equation}
\label{eq: translation optimization}
\bar{\mathbf{T}}_{opt} = \underset{^{\neg}\mathbf{T}}{\mathrm{argmin}} \sum_{i}{(F_{ICP}[\bar{\mathbf{T}}]+\lambda F_{CT}[\bar{\mathbf{T}}])},
\end{equation}
where the objective function comprises two components: $F_{ICP}[\cdot]$, which addresses geometric alignment based on the distance from points to a distribution. $F_{CT}[\cdot]$ represents the continuous-time-based translation constraint. $\lambda$ serves as an adjustable weight for $F_{CT}[\cdot]$. Given a correspondence $\left \langle\mathbf{p}_s, \mathbf{m}_k\right \rangle \in \mathcal{C}$, the $F_{ICP}[\cdot]$ is expressed by:
\begin{align}
F_{ICP}[\bar{\mathbf{T}}] & = \left \| N_j \cdot (\bar{\mathbf{p}}_k - \bar{\mathbf{T}} \mathbf{p}_s) \right \|_{\bm{\Omega}_{ICP}^{-1}}, \\
\bm{\Omega}_{ICP}  &= \bar{\bm{\Omega} }_k +\bar{\mathbf{T}} ^{\top}\bm{\Omega}_s \bar{\mathbf{T}},
\end{align}
where $N_j$ denotes the number of points contained in the voxel $\mathbf{m}_k$. $\bm{\Omega}$ represents the covariance matrix. 
Given the uniform motion model, the translation between two scans holds a linear relationship with time. To meet this assumption, the function $F_{CT}[\cdot]$, as it applies to each point $\mathbf{p}_i \in \mathcal{P}_t$, is expressed as follows:
\begin{align}
F_{CT}[\bar{\mathbf{T}}] & = \left \| \mathbf{t}^{n} - \mathbf{t}^{n-1} \right \|_{\bm{\Omega}_{CT}^{-1}},\\
\mathbf{t}^{n} &= (\bar{\mathbf{T}}\mathbf{p}_{i}- \breve{\mathbf{T}}[\bar{\mathbf{T}}_{flp}]^{-1} \mathbf{p} _{i}),\\ 
\bm{\Omega}_{CT} &= \mathbf{t}^{n} \otimes \mathbf{t}^{n}.
\end{align}
Here the translation element in $\bar{\mathbf{T}}_{flp}$ and rotation element in $\breve{\mathbf{T}}$ are the results estimated from the forward location prediction and rotation registration, respectively.
The covariance matrix $\bm{\Omega}_{CT}$, represented as the tensor product of $\mathbf{t}^{n}$ itself, where the larger element in $\mathbf{t}^{n}$ means the corresponding dimension suffers greater penalty in the optimization.
$F_{ICP}(\cdot)$ is utilized to achieve the geometric alignment of sensor data, whereas $F_{CT}(\cdot)$ ensures that the vehicle maintains as continuous and uniform a motion as possible.
The final translation transformation $\bar{\mathbf{T}}$ is a combination of $\bar{\mathbf{T}}_{flp}$ and $\bar{\mathbf{T}}_{opt}$, which is calculated by
\begin{equation}
    \bar{\mathbf{T}} = \bar{\mathbf{T}}_{opt} \cdot{\bar{\mathbf{T}}_{flp}}.
\end{equation}
It is important to note that solving the problem in Eq. \ref{eq: translation optimization} does not involve the matching process described in Sec. \ref{rotation estimation}; instead, we directly inherit $\mathcal{C}$ from the rotation registration. This approach is developed to accelerate the processing speed of the front-end.

\subsection{Back-End Mapping and Loop Closure}
\label{mapping and lc}
The back-end refines the transformation output from the front-end, facilitating the generation of high-quality global poses and environmental map. 
The back-end comprises two primary modules: local scan-to-submap alignment and global pose optimization. 
At the local level, the scan-to-submap alignment employs a precise registration method to align the latest scans with the accumulated local submaps, thereby achieving more accurate LiDAR odometry. 
At the global level, a factor graph is incrementally constructed from the accumulated keyframes, which adjusts the poses of each historical keyframe to minimize overall historical errors.

\subsubsection{Scan-to-Submap Optimization}

Firstly, the scan-to-submap alignment is utilized to further optimize the pose estimation of ground vehicles. 
The global point cloud map is constructed by historical keyframes and each keyframe 
$\mathbb{F}$ consisting of edge feature $\mathcal{F}_e$ and plane feature $\mathcal{F}_p$, formulated as $\mathbb{F} = \mathcal{F}_e, \mathcal{F}_p \}$. To reduce memory overhead, keyframes are periodically selected at a predefined time interval. A predefined number of keyframes within a sliding time window are selected to construct the submap $\mathcal{M}$, which is expressed by
\begin{align}
\mathcal{M} & = \{ \mathcal{M}_{e}, \mathcal{M}_{p} \}, \\
\mathcal{M}_{e} &= \{ \mathbb{F}^{i}_e, \mathbb{F}^{i-1}_e, \mathbb{F}^{i-2}_e, \dots, \mathbb{F}^{i-k+1}_e \}, \\
\mathcal{M}_{p} &= \{\mathbb{F}^{i}_p, \mathbb{F}^{i-1}_p,\mathbb{F}^{i-2}_p, \dots, \mathbb{F}^{i-k+1}_p \},
\end{align}
where $\mathcal{M}_{e}$ is the edge submap which is a set containing all the edge features in the keyframes. $\mathcal{M}_{p}$ is the planar submap. The scan-to-submap alignment is cast into an optimization problem, which can be written as
\begin{equation}
\mathbf{T}_b  = \underset{\mathbf{T}_b }{\mathrm{argmin}} \sum_{j} (F_{e}[\mathbf{T}_b \mathbf{p}_{e}] + F_{p}[\mathbf{T}_b \mathbf{p}_{p}]).
\end{equation}
The variable to be optimized is the transformation matrix $\mathbf{T}_b$ between the scan and the submap. $\mathbf{p}_{e}$ and $\mathbf{p}_{p}$ are points in $\mathcal{F}_{e}$ and $\mathcal{F}_{p}$. 
Here $F_{e}[\cdot]$ and $F_{p}[\cdot]$ are the cost function of edge features and planar features, respectively.  In our method, $F_e[\cdot]$ measures the distance between the current edge feature points $\{\mathbf{p}_{e} \in \mathcal{P}\}$ and the corresponding points in the submap $\{\mathbf{p}_{e}^{\prime} \in \mathcal{M}_e\}$. $F_p[\cdot]$ determines the distance between the current planar feature points $\{\mathbf{p}_{p} \in \mathcal{P}\}$ and the corresponding points in the submap $\{\mathbf{p}^{\prime}_{p} \in \mathcal{M}_p \}$. These distances are calculated as follows:
\begin{align}
F_e[\mathbf{T}_b \mathbf{p}_{e}] & =  \beta_e \frac{\left \| (\mathbf{p}^{\prime}_{e}-\mathbf{T} _{b}\mathbf{p}_{e}) \times (\mathbf{p}^{\prime}_{e}-\mathbf{T} _{b}\mathbf{p} _{e}-\mathbf{n}_e ) \right \| }{\left \| \mathbf{n}_e \right \|  }, \\
F_p[\mathbf{T}_b \mathbf{p}_{p}] & = \beta_p \frac{(\mathbf{T} _{b}\mathbf{p} _{p} - \mathbf{p}^{\prime}_{p}) \cdot \mathbf{n}_p}{\mathbf{n}_p },\\
\beta_e &= \sum \frac{\left \| \mathbf{n} _{e} \right \| }{\left \| (\mathbf{p} ^{\prime}_{e}-\bar{\mathbf{p} }^{\prime}_{e})\times \mathbf{n}_{e} \right \| },\\
\beta_p &= \frac{1}{\left |\sum \mathbf{n} ^{\top}_p \mathbf{p} ^{\prime}_{p} +1 \right |},
\end{align} 
where $\mathbf{n}_{e}$ is the unit vector of the edge feature while $\mathbf{n}_{p}$ is the normal vector of the planar feature. In addition, $\beta_{e}$ and $\beta_{p}$ serve as weight parameters for separate residuals, prioritizing features that exhibit smaller distances to their corresponding edges or planes.

\subsubsection{Global Optimization and Loop Closure}
\begin{figure}[t]
\centerline{\includegraphics[scale=0.6]{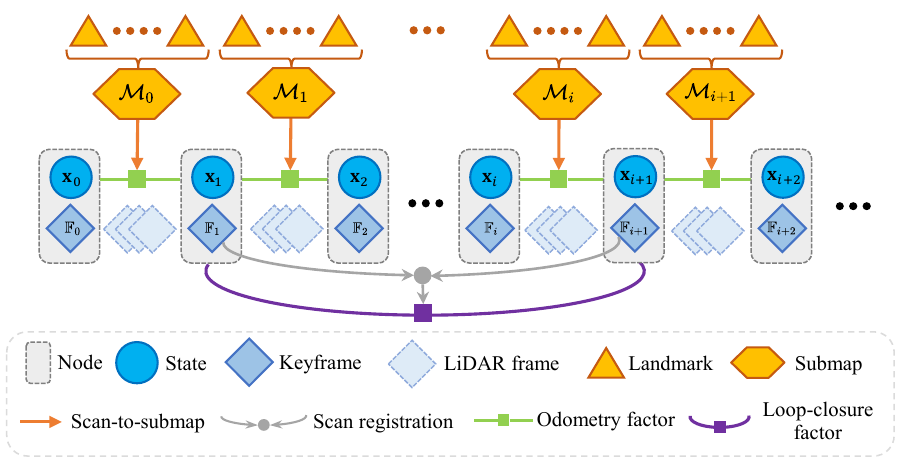}}
\caption{Factor graph structure of ROLO-SLAM. Two types of factor, incorporating the odometry factor and the loop-closure factor, are established as the vehicle moves.}
\label{fig: factor_graph}
\end{figure}

Pose optimization on the global scale is typically modeled as a Maximum A Posterior (MAP) Problem. ROLO-SLAM leverages the factor graph (FG) model to solve the MAP. The whole FG is composed of nodes and edges with different factors. Each node stores a state at that moment and we define the state as the vehicle pose in the world frame, i.e., $\mathbf{X}_i = [\mathbf{R}_i\ |\ \mathbf{t}_i]$.
Each $\mathbf{X}_i$ is associated with a keyframe $\mathbb{F}_{i}$. Furthermore, we define two factors: odometry factor and loop-closure factor. 
The overall factor graph structure is shown in Fig. \ref{fig: factor_graph}.
\begin{figure}[t]
\centerline{\includegraphics[scale=0.29]{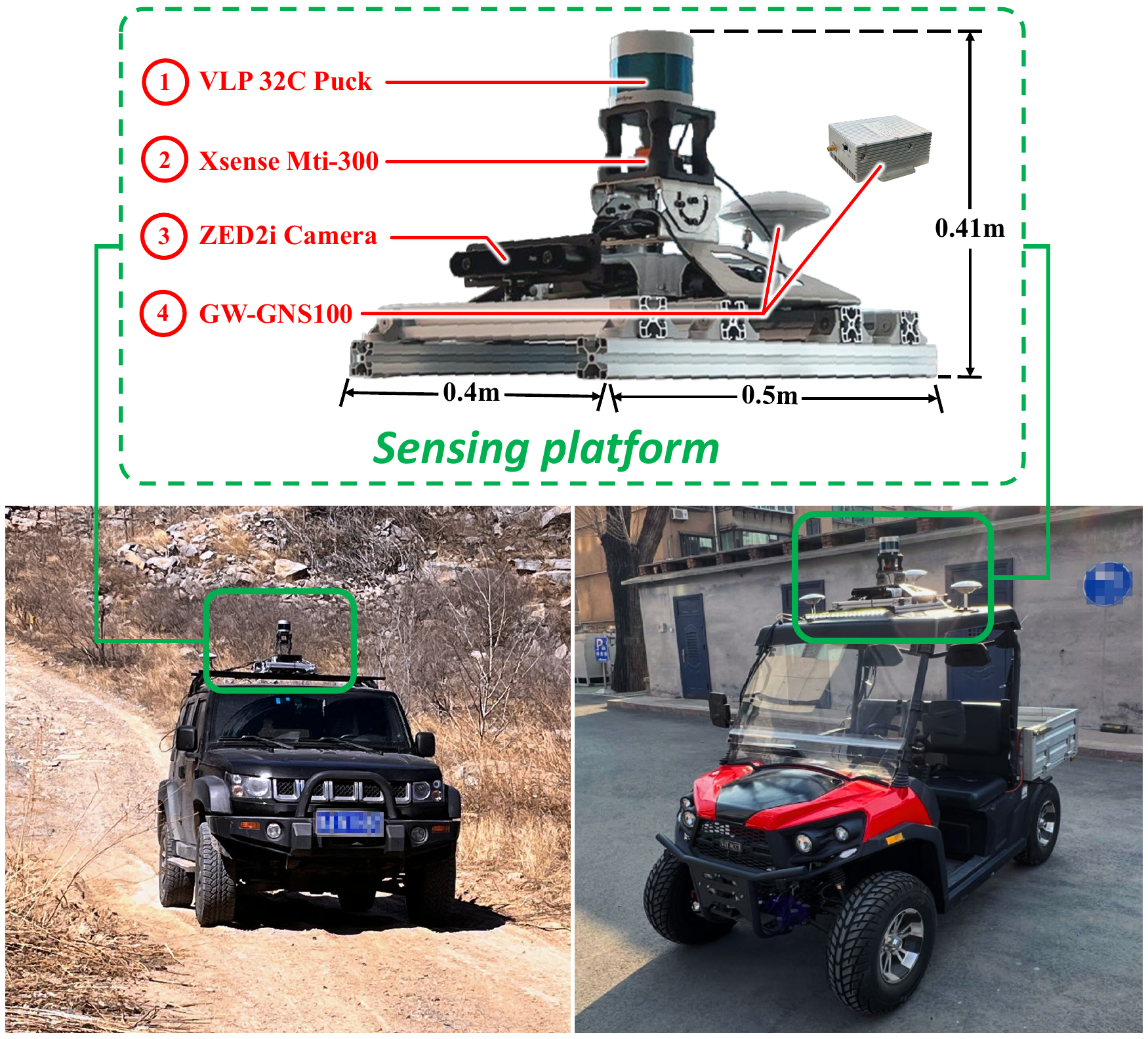}}
\caption{Ground vehicles with sensing platform in off-road and campus datasets}
\label{fig: platform}
\end{figure}
\begin{table*}[t]
\centering
\caption{Dataset parameter setting}
\label{tab: dataset}
\begin{tabular}{ccccccccc}
\hline
\multicolumn{1}{c|}{Dataset} & \multicolumn{1}{c|}{\begin{tabular}[c]{@{}c@{}}KITTI\_raw\\ seq. 00\end{tabular}} & \begin{tabular}[c]{@{}c@{}}KITTI\_raw\\ seq. 05\end{tabular} & \begin{tabular}[c]{@{}c@{}}KITTI\_raw\\ seq. 08\end{tabular} & \begin{tabular}[c]{@{}c@{}}SDU Campus\\ Qianfo\end{tabular} & \begin{tabular}[c]{@{}c@{}}SDU Campus\\ Xinglong\end{tabular} & Offroad1 & Offroad2 & Offroad3 \\ \hline
Have Loop                    & \checkmark                                                         & \checkmark                                    & \checkmark                                       & \checkmark                                         &      &      & &\checkmark     \\
Data Frames                  &4541                                                                                   &2761                                                              &4071                                                                 &5370                                                                   &5201      &1882      &2339 &  4336   \\
Max Ground Height (m)           &3.225                                                                                  &6.573                                                              &5.051                                                                 &33.389                                                                   &34.285      &0.797      &0.050 &23.705     \\
Min Ground Height (m)           &-22.295                                                                                   &-12.092                                                              &-40.681                                                                 &-0.105                                                                   &-0.301      &-29.308      &-31.958   &-2.986   \\
Trajectory Length (m)            &3387.473                                                                                   &1997.451                                                              &2879.163                                                                 &2578.197                                                                   &3580.265      &927.516      &1402.278  &1140.229    \\
Time Duration (s)               &470                                                                                   &287                                                              &422                                                                 &541                                                                   &521      &188      &234  &434   \\ \hline
\end{tabular}
\end{table*}
\begin{figure*}[t]
	\setlength{\belowcaptionskip}{-10pt}
	\centering
	\includegraphics[scale=0.395]{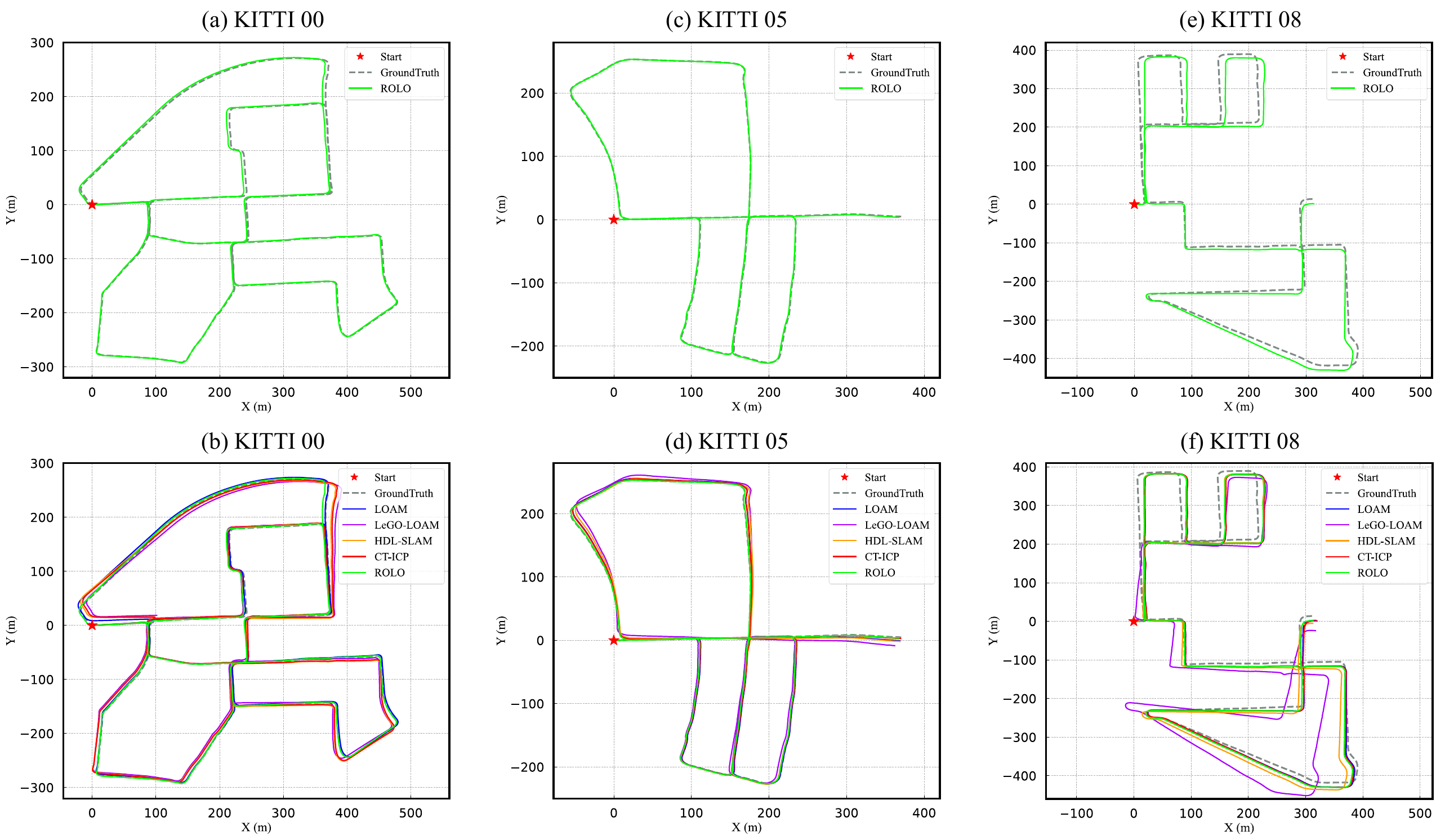}
        \vspace{-0.2cm}
	\caption{The trajectory estimation of \textbf{ROLO} and other odometry frameworks for \textit{KITTI-odometry} dataset \textit{sequence \texttt{00}}, \textit{\texttt{05}} and \textit{\texttt{08}}.}
	\label{fig: kitti_trajectory}
\end{figure*}
\begin{figure*}[t]
	\setlength{\belowcaptionskip}{-10pt}
	\centering
        \includegraphics[scale=0.31]{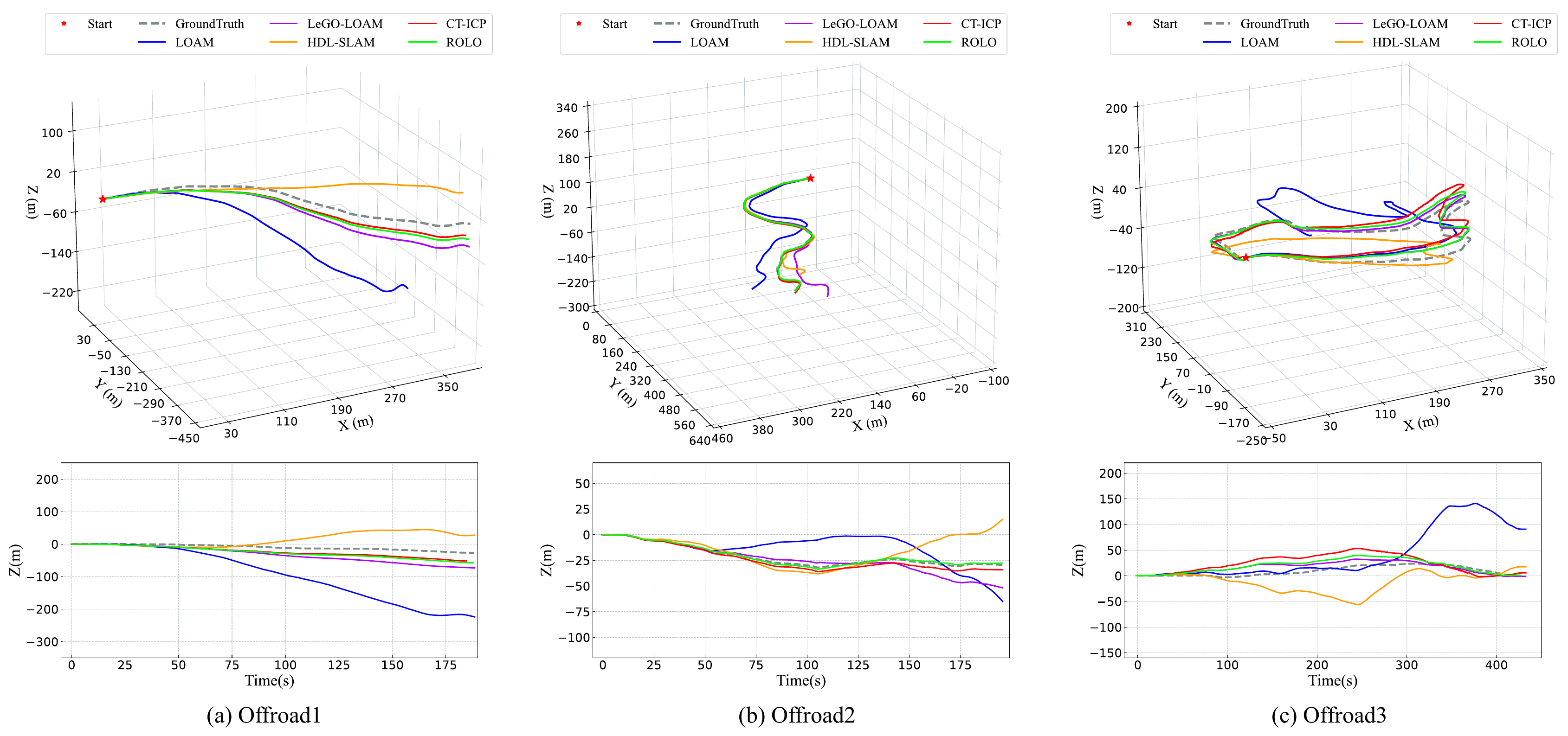}
        \vspace{-0.2cm}
	\caption{The trajectory estimation of our method and alternative methods for the off-road datasets (\textit{Offroad1}, \textit{Offroad2}, \textit{Offroad3}).}
    \label{fig: off_trajectory}
\end{figure*}
The odometry factor constrains the transformation between adjacent states, analogous to a Markov chain. The transformation of adjacent nodes is given by the scan-to-submap alignment. This factor rejects outliers of state estimated by the scan-to-submap alignment and smooths the locomotion trajectory.
The loop-closure factor is used to address accumulated errors in long-term and large-scale scenarios. To build this factor, we establish a stable search window $\mathcal{B}$, centered on the current state $\mathbf{X}_c$. There are states and keyframes in the window $\mathcal{B}$, denoted as $\mathcal{B} = \{\mathbb{F}_{0}, \mathbb{F}_{1}, \dots\, \mathbb{F}_{i}, \dots\}$. 
During the vehicle moving, a checking thread is consistently executed to assess the similarity between each $\mathbb{F}_{i} \in \mathcal{B}$ and the current keyframe $\mathbb{F}_{c}$. Upon identifying significant similarity between $\mathbb{F}_{i}$ and $\mathbb{F}_{c}$, a feature registration is employed to determine the transformation $\stackrel\frown{\mathbf{T}} _{\mathbb{F}_{c},\mathbb{F}_{i}}$ of $\mathbb{F}_{i}$ and $\mathbb{F}_{c}$. Then, the loop-closure tuple $\mathcal{L}_{c,i}$ is established, denoted as:
\begin{equation}
\mathcal{L} _{c, i} = \left \langle \mathbf{X} _c, \mathbf{X} _i, \mathbf{\stackrel\frown{T} } _{\mathbb{F}_{c}, \mathbb{F} _{i}} \right \rangle .
\end{equation}
These loop-closure tuples are transformed into loop-closure factors in the FG, establishing constraints between nodes introduced at different time instances. 
By employing the FG for global pose optimization, the vehicle's pose is refined,   enabling adaptive adjustments of all nodes to minimize global discrepancies to effectively eliminate accumulated errors.

\section{Experimental Evaluation}
\label{experiments}
\subsection{Platform and Experiments Setting}
To evaluate the performance of the ROLO-SLAM, comprehensive experiments focusing on pose estimation accuracy, robustness, computational efficiency and mapping results are performed using both public datasets and our elaborate datasets. The following illustrates the details of these datasets.

\begin{itemize}
    \item \textit{KITTI-odometry (sequence \texttt{00}, \texttt{05} and \texttt{08})} \cite{Geiger2012kitti}, is a benchmark dataset for odometry and perception tasks, which is captured in urban, rural, and highway scenarios. The odometry benchmark offers multi-model data including LiDAR point clouds and camera images in grayscale and color. In addition, KITTI maintains the calibration parameters and benchmark results. We use \textit{KITTI-odometry} dataset to evaluate the localization accuracy in the horizon direction.
    \item \textit{Real Off-Road scene (Offroad1, Offroad2, Offroad3)}. These datasets are collected in off-road environments on a mountain around our campus. They include various scenarios, like steep slopes, muddy roads, grasslands, and hard-surfaced roads. Furthermore, the dataset offers multiple sensor data and ground truth generated by the multi-sensor-based SLAM method. These datasets are suitable for evaluating 6D pose estimation.
    \item \textit{SDU Campus Scene (Qianfo campus and Xinglong campus)}. We elaborate the SDU datasets in Qianfo campus and Xinglong campus of our university, which both incorporate variant terrains, like slopes and uneven floors. The ground truth of localization is offered by the fusion of LiDARs, IMU, and GPS sensors. This dataset is leveraged to evaluate the overall performance of the developed method.
\end{itemize}

Our established datasets are recorded by two ground vehicles as shown in Fig. \ref{fig: platform}. To collect information in outdoor environments, these vehicles are equipped with the same sensing platform. The sensing platform has multiple sensors including a VLP-32C Puck 3D LiDAR, a ZED2i stereo camera, a GW-GNS100 GNSS system, and a Mti-300 9-axis IMU. The ground truth of \textit{Off-road} and \textit{SDU Campus} datasets are generated by the multi-sensor-based SLAM method \cite{liu2023glio}. To calibrate heterogeneous sensors, we utilize offline calibration methods to obtain intrinsic and extrinsic parameters. Specifically, we first use the Kalibr toolbox \cite{rehder2016kalibr} to calibrate the intrinsic parameters and relative transformation between IMU and stereo camera. Subsequently, we performed the LiDAR camera calibration using the Autoware toolkit \cite{kato2018autoware} to obtain the relative transformation.
For more details, Tab. \ref{tab: dataset} lists parameters of all these datasets used in the evaluation.
For the sake of brevity, in the experimental demonstration, we denote \textbf{ROLO} as ROLO-SLAM. To highlight the performance of the our method, we conduct abundant comparison experiments with the following state-of-the-art methods:
\begin{itemize}
    \item \textbf{LOAM} \cite{2014loam}, firstly leverages the simple but efficient registration to achieve precise LiDAR odometry tasks and indicates high performance in durational localization accuracy.
    \item \textbf{LeGO-LOAM} \cite{lego_loam}, towards vertical accumulated drift problem, segments the ground observation and divides the 6-DOF pose estimation into two separate steps. It exhibits excellent accuracy in uneven terrains with ground vehicles.
    \item \textbf{CT-ICP} \cite{dellenbach2022cticp} presents a continuous time ICP method to achieve smoother motion trajectories by incorporating time and velocity constraints. This approach demonstrates enhanced performance, particularly in the context of continuous motion scenarios.
    \item \textbf{HDL-SLAM} \cite{koide2019hdl_slam} integrates a variety of cutting-edge registration and loop-closure techniques to achieve localization. Furthermore, it incorporates people behavior prediction and ground perception capabilities, thereby showcasing superior performance in long-term localization tasks.
\end{itemize}

The implementation for ROLO-SLAM system is on Robot Operation System (ROS) Noetic, Ubuntu 20.04. All evaluated algorithms are executed on a Lenovo Y9000P Laptop with an Intel Core i7th CPU, an Nvidia RTX2070 GPU and 32 GB RAM.

\subsection{Localization Accuracy Evaluation}
\begin{figure*}[t]
	\centering
	\includegraphics[scale=0.39]{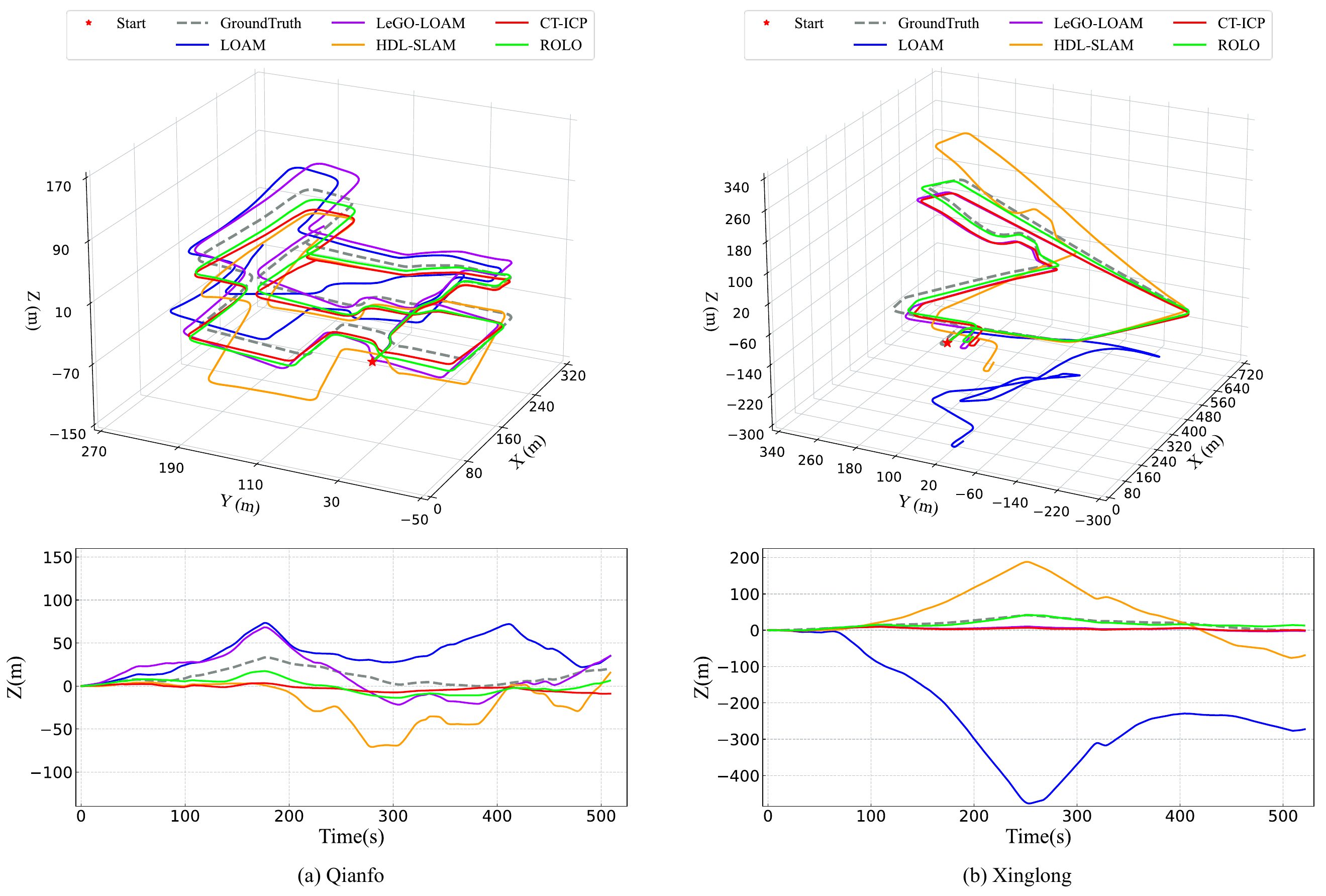}
	\caption{The trajectory estimation of our method and alternative methods for the SDU campus datasets (\textit{Qianfo}, \textit{Xinglong}).}
	\label{fig: sdu_trajectory}
\end{figure*}
In this section, we evaluate the localization accuracy of our method via comparing with different approaches. In \textit{KITTI-odometry} datasets (\textit{seq. \texttt{00}}, \textit{\texttt{05}} and \textit{\texttt{08}}), the trajectory results are shown in Fig. \ref{fig: kitti_trajectory}. 
The trajectory of \textbf{ROLO} aligns closely with the ground truth (GT) in these three datasets. Specifically, \textbf{ROLO} almostly overlaps with GT, especially in Fig. \ref{fig: kitti_trajectory}(c). \textbf{LOAM} and \textbf{CT-ICP} also have high trajectory similarity with GT. \textbf{LOAM} has the convincing performance as highlighted in Fig. \ref{fig: kitti_trajectory}(b). This is caused by rich features in urban scenes, which offer a registration basis for these two methods. Instead, other methods showcase different biases of varying magnitudes. 
\textbf{LeGO-LOAM} and \textbf{HDL-SLAM} exhibit obvious trajectory errors in Fig. \ref{fig: kitti_trajectory}(f), though approximately trace correct trajectories in the \textit{KITTI seq. 00} and \textit{05}. Compared with alternative approaches, \textbf{ROLO} exhibits the greatest degree of alignment with the GT. These results highlight the exceptional localization performance of our method in the horizon direction.

\begin{table*}[t]
\centering
\renewcommand\arraystretch{1.3}
\caption{RMSE (translation (m)/rotation (deg)) of methods with ground truth}
\label{tab: RMSE}
\begin{tabular}{lcccccccc}
\hline
            & KITTI seq.00 & KITTI seq.05 & KITTI seq.08 & \begin{tabular}[c]{@{}c@{}}SDU camp\\ (Qianfo)\end{tabular} & \begin{tabular}[c]{@{}c@{}}SDU camp\\ (Xinglong)\end{tabular} & Offroad1 & Offroad2 & Offroad3 \\ \hline
ROLO (ours) & 0.200/2.369  & \textbf{0.085}/\textbf{2.729}  & \textbf{0.291}/3.045  & \textbf{0.446}/\textbf{0.149 }                                                & \textbf{0.348}/0.046                                                   & \textbf{0.524}/\textbf{0.552} & \textbf{0.056}/\textbf{0.035} & \textbf{0.269}/\textbf{0.21}  \\
LOAM        & \textbf{0.189}/2.327  & 0.100/2.94   & 0.294/3.115  & 1.269/0.342                                                 & 6.991/0.853                                                   & 7.399/6.202 & 1.953/1.044 & 2.747/1.016 \\
LeGO-LOAM   & 0.665/\textbf{2.239}  & 0.320/2.865  & 0.977/\textbf{3.008}  & 0.591/0.304                                                 & 0.571/\textbf{0.045}                                                   & 1.354/5.43  & 1.074/0.397 & 0.284/0.217 \\
CT-ICP      & 0.245/2.245  & 0.092/2.841  & 0.306/3.023  & 0.538/0.18                                                  & 0.590/0.052                                                   & 0.765/5.462 & 0.149/0.137 & 0.539/0.297 \\
HDL-SLAM    & 0.228/2.245  & 0.132/2.844  & 0.449/3.028  & 1.087/0.573                                                 & 1.774/0.315                                                   & 1.967/5.591 & 0.691/0.465 & 0.982/0.473 \\ \hline
\end{tabular}
\end{table*}

To evaluate the efficacy of localization in the vertical direction, we compare our method with others using \textit{Offroad1}, \textit{Offroad2} and \textit{Offroad3} datasets. Note that these datasets are characterized by numerous steep slopes and uneven road surfaces. Therefore, these environmental features lead to intensive changes in the position and attitude of the vehicle. The trajectory results and corresponding elevation curves are depicted in Fig. \ref{fig: off_trajectory}(a)-(c).
\begin{figure*}[t] 
    \setlength{\abovecaptionskip}{-5pt}
    \setlength{\belowcaptionskip}{-10pt}
    \centering
	\includegraphics[scale=1.01]{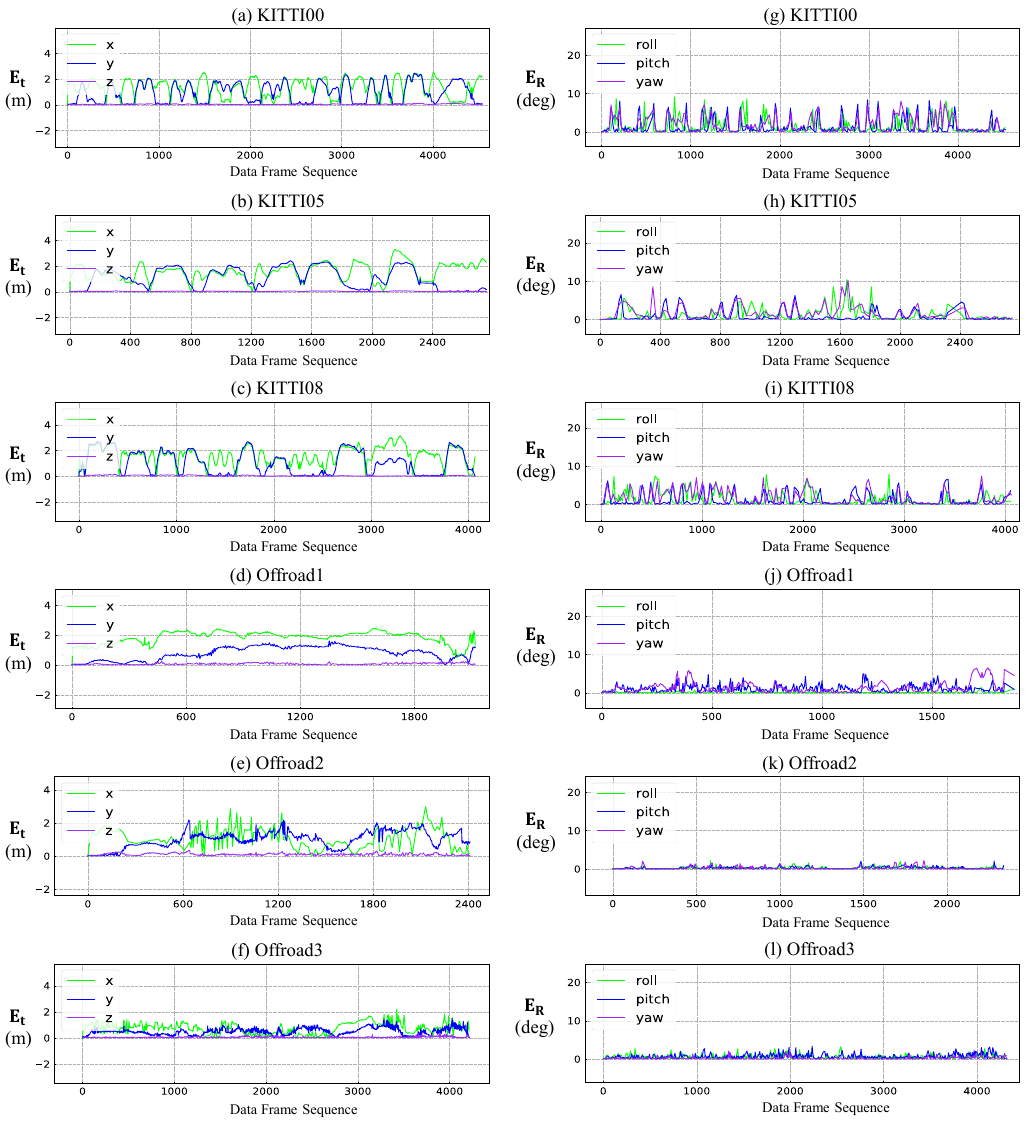}
    \vspace{0.2cm}
	\caption{ $\mathcal{\mathbf{E}}^{\mathbf{R} }$ and $\mathcal{\mathbf{E}}^{\mathbf{t} }$ curves of \textbf{ROLO} in \textit{KITTI} and \textit{Off-road} datasets. The each dimension of $\mathcal{\mathbf{E}}^{\mathbf{R} }$ and $\mathcal{\mathbf{E}}^{\mathbf{t} }$ is separately depicted in subfigures.} 
	\label{fig: error_curve}
\end{figure*}

In comparison to other approaches, the 3D trajectory of our method exhibits the highest similarity with the GT trajectory, particularly in Fig. \ref{fig: off_trajectory}(b). In addition, the elevation curve of our method closely fits with the GT as shown in Fig. \ref{fig: off_trajectory}(a)-(b). Notably, the elevation curve and trajectory of \textbf{LeGO-LOAM} showcase a promising performance in Fig.\ref{fig: off_trajectory}(c).
This implies that ground surface segmentation offers positive guidance in the estimation of roll and pitch angle.
In contrast, other methods utilize conventional point-to-plane and point-to-line registration techniques, which prove to be less effective in scenarios with sparse features and drastic changes in surface geometry.

The performance of pose estimation with these approaches is further validated using \textit{SDU campus} datasets. The results are shown in Fig. \ref{fig: sdu_trajectory}(a)-(b). 
\begin{table*}[t]
\centering
\renewcommand\arraystretch{1.3}
\caption{RMSE (translation (m)/rotation (deg)) frame-to-frame registration of ours and traditional registration methods}
\label{tab: iRMSE}
\begin{tabular}{lcccccccc}
\hline
            & KITTI seq.00 & KITTI seq.05 & KITTI seq.08 & \begin{tabular}[c]{@{}c@{}}SDU camp\\ (Qianfo)\end{tabular} & \begin{tabular}[c]{@{}c@{}}SDU camp\\ (Xinglong)\end{tabular} & Offroad1 & Offroad2 & Offroad3 \\ \hline
Ours    & 1.603/2.246  & 0.458/2.844  & \textbf{0.876}/\textbf{2.586}  & \textbf{0.664}/\textbf{0.223}                                                 & \textbf{1.547}/\textbf{0.321}                                                   & 1.513/0.334 & \textbf{1.864}/\textbf{0.364} & \textbf{0.993}/\textbf{0.397} \\
ICP     & 4.321/2.728  & 3.866/3.164  & 4.790/2.71   & 2.159/1.017                                                 & 7.116/0.831                                                  & \textbf{1.080}/\textbf{0.326} & 2.539/0.655 & 5.144/1.52  \\
NDT     & 4.488/3.075  & 3.930/3.937  & 4.869/3.453  & 2.610/1.053                                                 & 6.157/1.42                                                    & 6.946/1.054 & 6.977/1.735 & 2.699/1.726 \\
PTP ICP & \textbf{0.61}/\textbf{2.245}  & \textbf{0.415}/\textbf{2.837}  & 1.156/2.99   & 1.395/0.426                                                 & 3.845/1.279                                                   & 1.962/1.87  & 4.503/0.946 & 1.167/0.814 \\ \hline

\end{tabular}
\end{table*}
\begin{table}[t]
\centering
\renewcommand\arraystretch{1.3}
\tabcolsep=0.3cm
\caption{Absolute error for inter-frame (\textit{mean}/\textit{std.}) towards \textit{SDU Campus} datasets}
\label{tab: sdu_error}
\begin{tabular}{cccc}
\hline
Datasets                 & Methods   & \multicolumn{1}{c}{$\mathcal{\mathbf{E}}^{\mathbf{t} }$ (m)} & \multicolumn{1}{c}{$\mathcal{\mathbf{E}}^{\mathbf{R} }$ (deg)} \\ \hline
\multirow{5}{*}{Qianfo}   & ROLO      & \textbf{0.057}/\textbf{0.042}                                 & \textbf{0.134}/0.219                                \\
                          & LOAM      & 0.108/0.099                                 & 0.286/0.26                               \\
                          & LeGO-LOAM & 0.107/0.096                                 & 0.249/0.333                                \\
                          & CT-ICP    & 0.084/0.048                                  & 0.18/\textbf{0.124}                                \\
                          & HDL-SLAM  & 0.112/0.152                                 & 0.266/0.603                                \\ \hline
\multirow{5}{*}{Xinglong} & ROLO      & \textbf{0.013}/0.072                                 & \textbf{0.073}/0.211                                \\
                          & LOAM      & 0.304/0.222                                 & 0.245/0.505                                \\
                          & LeGO-LOAM & 0.074/0.103                                 & 0.152/0.217                                \\
                          & CT-ICP    & 0.043/\textbf{0.061}                                 & 0.08/\textbf{0.142}                                \\
                          & HDL-SLAM  & 0.106/0.107                                 & 0.184/0.241                                \\ \hline
\end{tabular}
\end{table}

The trajectory of our method maintains close with GT in Fig. \ref{fig: sdu_trajectory}(a). In Fig. \ref{fig: sdu_trajectory}(b), the trajectory and elevation curve of our method are basically overlapped with GT, which emphasizes the high performance of \textbf{ROLO} in localization accuracy.
\textbf{CT-ICP} and \textbf{LeGO-LOAM} yield impressive results only in Fig. \ref{fig: sdu_trajectory}(b). However, other methods exhibit significant pose drift, especially in terms of elevation positioning. The elevation curve of \textbf{LOAM} shows an opposite trend to that of GT in Fig. \ref{fig: sdu_trajectory}(b). These outcomes demonstrate substantial advantage of our method in 6-DOF pose estimation. 

Furthermore, Tab. \ref{tab: RMSE} lists the root mean square error (RMSE) of translation and rotation for all methods. Results show that our method has the lowest RMSE in the \textit{SDU} and \textit{Off-road} datasets. In the \textit{KITTI} dataset, \textbf{LOAM} and \textbf{LeGO-LOAM} also have low RMSE of localization, which showcases excellent performance in localization accuracy, benefiting from rich environmental features and ground segmentation. However, other approaches exhibit varying errors across different scenarios. \textbf{LOAM} and \textbf{HDL-SLAM} methods suffer from higher RMSE in \textit{SDU campus} and \textit{Off-road} datasets, which reflects significant drifts on uneven roads. 

In order to indicate the benefit of the proposed registration method in front-end, we compare our registration method with the conventional point cloud registration methods in the evaluated datasets. These conventional registration methods comprise ICP, NDT and point-to-plane ICP (PTP ICP) registration methods. We utilize the RMSE of frame-to-frame registration to assess the registration accuracy. The quantitive results are shown in Tab. \ref{tab: iRMSE}. Our method has ever minimal RMSE in \textit{SDU} and \textit{Off-road} datasets, though it cannot have the best performance in \textit{KITTI} and \textit{Offroad1} datasets. The results indicate that PTP ICP has more advantages to achieve registration relying on rich environmental features in urban scenes. Our front-end registration method can offer relatively accurate pose estimation to the back-end in uneven scenes.
\begin{figure*}[p]
\centering
\includegraphics[scale=0.43]{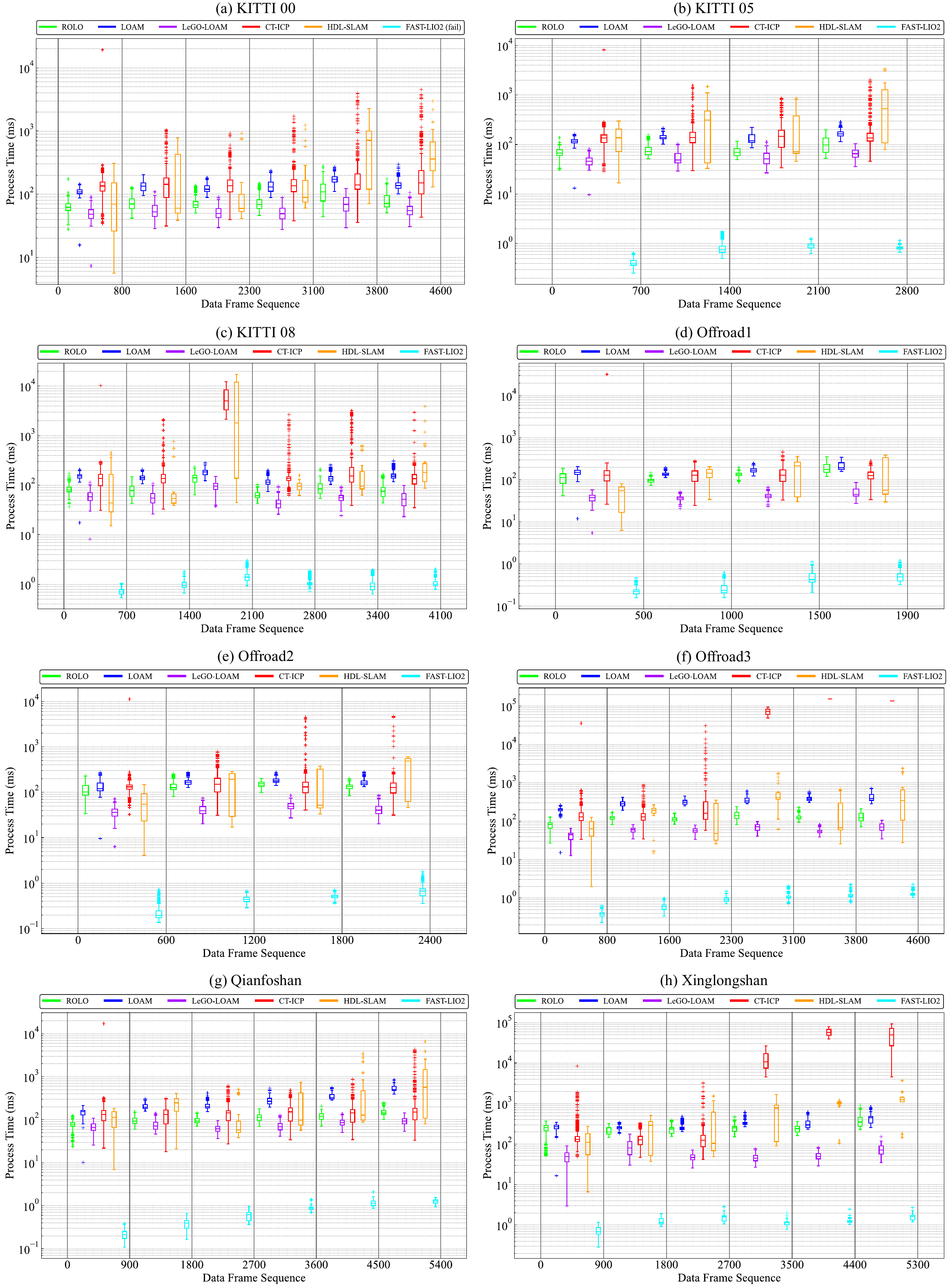}
\caption{Box plots of processing time per scan for all methods. Each dataset is divided into several pieces with respect to the data frames.}
\label{fig: time_box}
\end{figure*}
\begin{figure}[t]
\centerline{\includegraphics[scale=0.4]{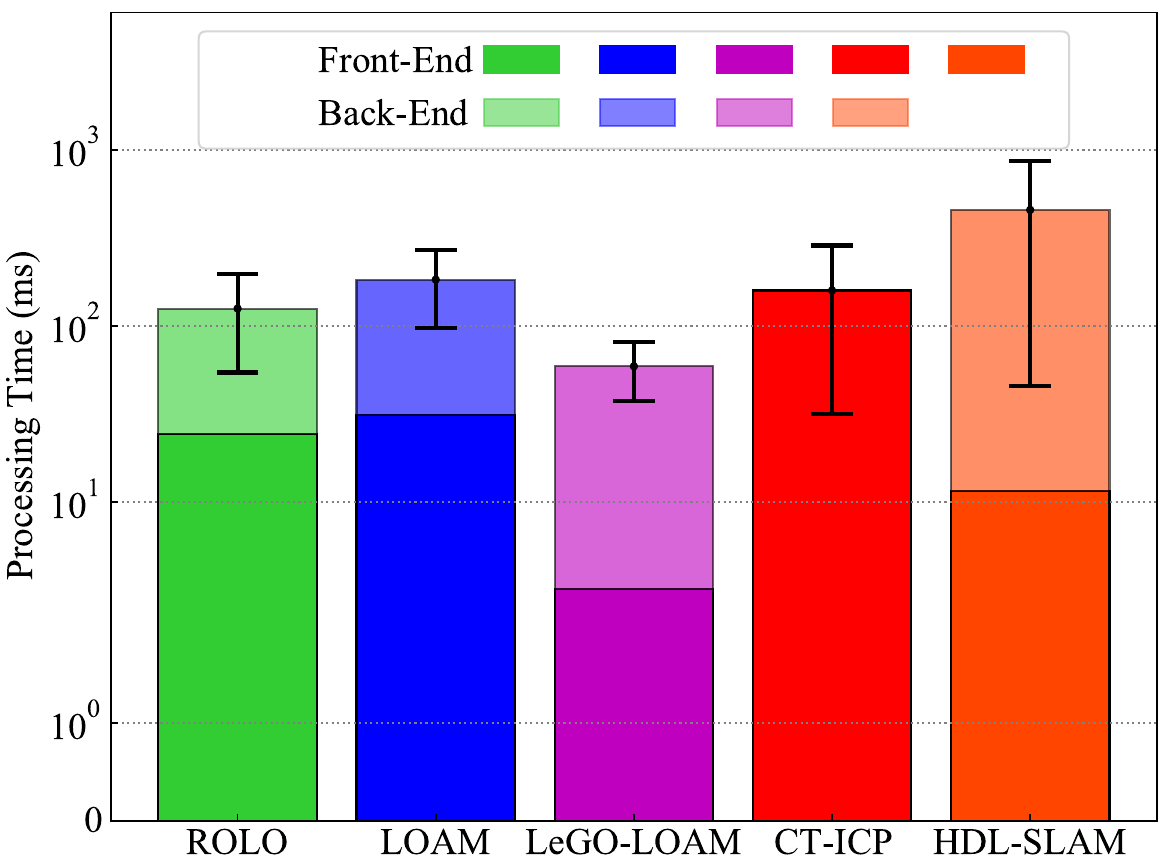}}
\caption{The overall processing time per scan for methods. Note that CT-ICP only has the processing time of front-end.}
\label{fig: total_time}
\end{figure}
\subsection{Robustness Evaluation}
Localization robustness reflects the capacity to sustain self-location stability in complex scenarios. 
To evaluate the robustness of localization, we utilize two absolute errors for inter-frame transformation to measure the localization robustness, here these two absolute errors are calculated as:
\begin{align}
\label{eq: absolute error_t}
\mathcal{\mathbf{E}}^{\mathbf{R}}_i & = \left | \mathbf{euler}(\mathbf{R}_{i})- \mathbf{euler}(\mathbf{R}^{gt}_{i}) \right|, \\
\label{eq: absolute error_r}
\mathcal{\mathbf{E}}^{\mathbf{t}}_i & = \left |\mathbf{t}_{i} - \mathbf{t}^{gt}_{i}\right |,
\end{align}
where for consecutive LiDAR scans $\mathcal{P}_i$ and $\mathcal{P}_{i+1}$, $\mathbf{R}_{i}$ and $\mathbf{t}_{i}$ are the rotation and translation between these two scans.
$\mathbf{R}^{gt}_{i}$ and $\mathbf{t}^{gt}_{i}$ are the rotation and translation of GT. In addition, $\mathbf{euler}(\cdot)$ calculates the Euler angles. $\mathcal{\mathbf{E}}^{\mathbf{R}}_i$ and $\mathcal{\mathbf{E}}^{\mathbf{t}}_i$ are both 3D vectors. They reflect the instantaneous accuracy of each frame-to-frame transformation. 

In these experiments, all results of $\mathcal{\mathbf{E}}^{\mathbf{R}}$ and $\mathcal{\mathbf{E}}^{\mathbf{t}}$ are obtained by executing different methods for 5 times in the evaluated datasets. 
Fig. \ref{fig: error_curve}(a)-(l) shows the $\mathcal{\mathbf{E}}^{\mathbf{R}}$ and $\mathcal{\mathbf{E}}^{\mathbf{t}}$ curves of \textbf{ROLO} in \textit{KITTI} and \textit{Off-road} datasets.
From a macroscopic perspective, our method has small amplitude and mean on datasets, which is particularly intuitive in Fig. \ref{fig: error_curve}(j)-(l). It is worth noting that the vertical errors using our method always retain a value close to zero as shown in Fig. \ref{fig: error_curve}(a)-(f).
Moreover, In such off-road scenes, \textbf{ROLO} has smaller amplitudes than the urban cases as shown in Fig. \ref{fig: error_curve}(d)-(f) and Fig. \ref{fig: error_curve}(j)-(l). These results indicate that our method maintains the stability of localization along with high instantaneous accuracy in off-road cases.  This benefits from the fact that \textbf{ROLO} achieves precise pose estimation by independently estimating rotation and translation at the front-end. Moreover,  the pose estimation is further refined by the back-end. 

To compare the localization robustness with other methods, we collect the values of $\mathcal{\mathbf{E}}^{\mathbf{R}}$ and $\mathcal{\mathbf{E}}^{\mathbf{t}}$ in \textit{SDU campus} datasets. Tab. \ref{tab: sdu_error} shows the \textit{mean} and standard covariance (\textit{std.}) of $\mathcal{\mathbf{E}}^{\mathbf{R}}$ and $\mathcal{\mathbf{E}}^{\mathbf{t}}$. Compared with other methods, \textbf{ROLO} has smaller \textit{std.} value, which guarantees the estimation stability of consecutive scans.
Of note is that \textbf{ROLO} has the smallest mean values. 
It demonstrates the estimation of our method is significantly precise during execution. Comparatively, \textbf{LOAM} and \textbf{HDL-SLAM} have higher mean values and \textit{std.} values, thereby it is performed inadequately in localization robustness.

\subsection{Computation Efficiency Evaluation}
To demonstrate that \textbf{ROLO} not only has excellent localization accuracy but also has high computation efficiency, we compare the computational time among these methods. We record the time for processing two scan frames with all these methods. 
We run every method for 3 times on each dataset. 

Fig. \ref{fig: time_box} showcase box plots of processing time per scan, whose data is from the total processing time including the front-end and the back-end.
To ultimately avoid the influence of the processing time from the increasing global map size,
we divide the dataset into several bins and evaluate the performance of different methods in each unit.
Note that \textbf{CT-ICP} is recorded by the whole processing times since it lacks back-end module. 
The height of boxes in Fig. \ref{fig: time_box} represents the variation range of the processing time. One can see that \textbf{ROLO} and \textbf{LeGO-LOAM} have stable variation range and means values over multiple bins as the accumulation of data frames, as shown in Fig. \ref{fig: time_box}(a)-(h). It means that \textbf{ROLO} and \textbf{LeGO-LOAM} have stable processing capacity, whose processing time is at around 100 ms.
However, the time boxes of \textbf{CT-ICP} are located at the high position in Fig. \ref{fig: time_box}(f), \ref{fig: time_box}(h), and have many outliers in Fig. \ref{fig: time_box}(a)-(c), \ref{fig: time_box}(f)-(h). 
It indicates that the additional continuous-time constraints without initial value may bring processing delay and uncertainty though the localization accuracy obtains improvement.
\textbf{HDL-SLAM} has the largest variation range in Fig. \ref{fig: time_box}(a)-(g), which reflects unstable processing capacity of \textbf{HDL-SLAM}.
In summary, \textbf{ROLO} has higher performance in terms of processing robustness and efficiency.

Fig. \ref{fig: total_time} shows the average processing times per scan in all datasets, and each column is stacked with processing times of the front-end and back-end, respectively. The error bar represents the variation range of total processing time.
The results show that the front-end of \textbf{ROLO} maintains fast processing speed, although it is divided into three successive modules. The total processing time is around 100 ms per scan, which satisfies the real-time requirement for robot applications.
\textbf{LeGO-LOAM} has the shortest processing time of the front-end but the performance is unsatisfactory in aspect of localization accuracy. Moreover, \textbf{CT-ICP} and \textbf{HDL-SLAM} have longer error bars, which means their processing efficiency is unreliable. \textbf{ROLO} is faster than \textbf{CT-ICP} and \textbf{HDL-SLAM} by at least $10\%$, which demonstrates the efficiency of the developed method. 
\begin{figure}[t]
\centerline{\includegraphics[scale=0.3]{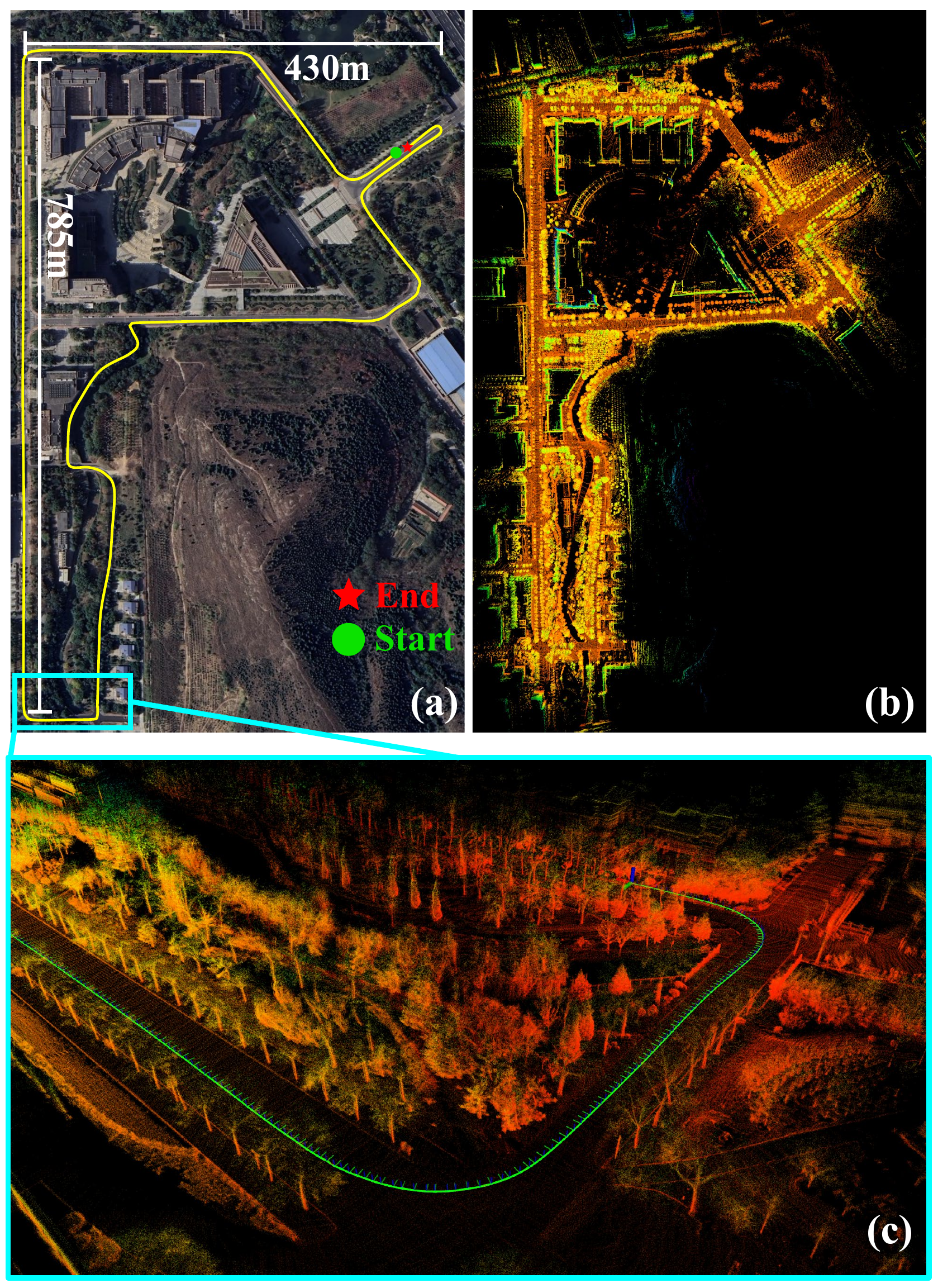}}
\caption{Mapping result for \textit{Xinglong} dataset. (a) shows the Google map, which indicates the trajectory and distance metrics. (b) displays overall point cloud map and (c) is captured at a specific scenario.}
\label{fig: xls_mapping}
\end{figure}
\begin{figure*}[t]
\centering
\includegraphics[scale=0.295]{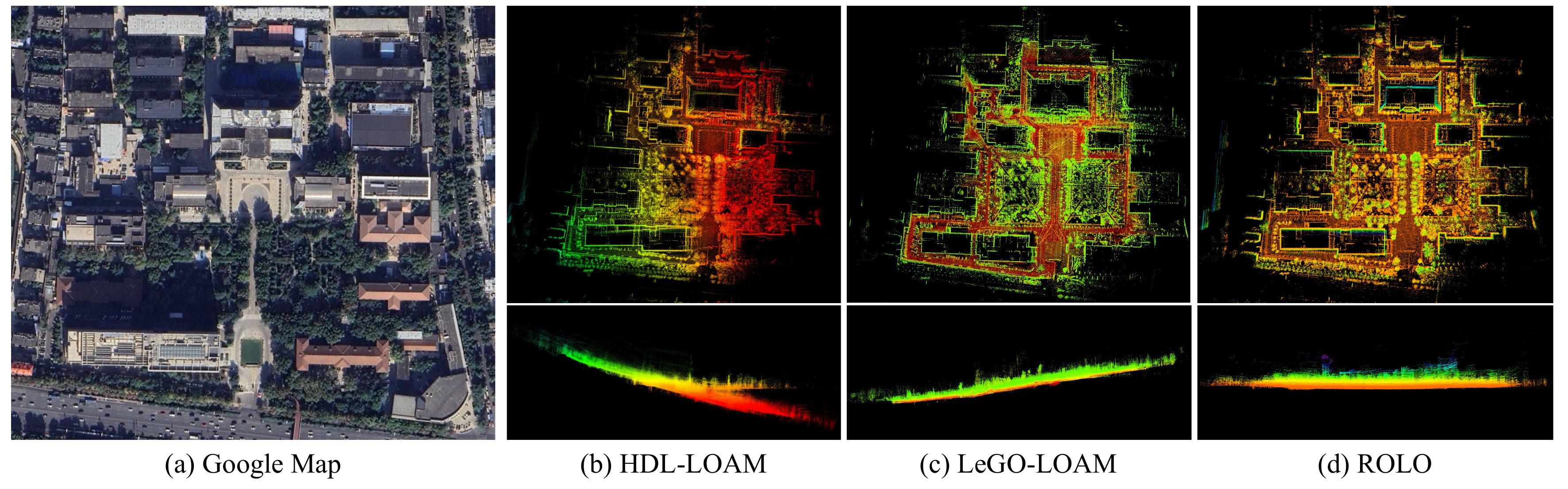}
\caption{Mapping results of \textbf{HDL-LOAM}, \textbf{LeGO-LOAM} and \textbf{ROLO} in \textit{Qianfo} dataset. In (b)-(d), the top figures show the overall maps while the bottom figures are captured at the side views.}
\label{fig: qfs_mapping}
\end{figure*}

\begin{figure}[t]
\centerline{\includegraphics[scale=0.15]{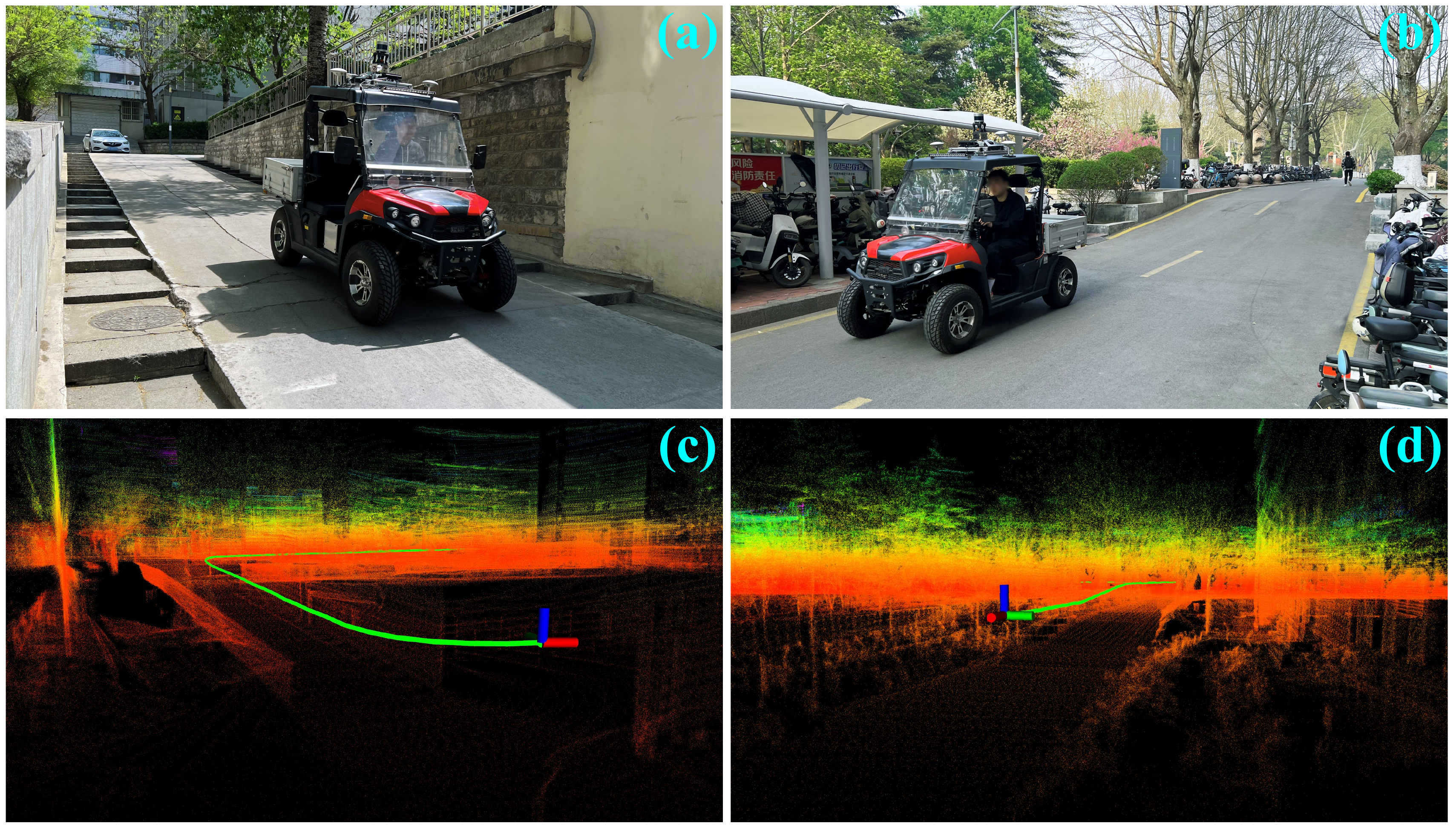}}
\caption{Mapping results in real slope cases of \textit{Qianfo} dataset. (a)-(b) show the real scenes with ground vehicles. (c)-(d) are the corresponding point cloud maps and the navigation trajectories of \textbf{ROLO}.}
\label{fig: two_slope}
\end{figure}

\begin{figure}[h]
\centerline{\includegraphics[scale=0.23]{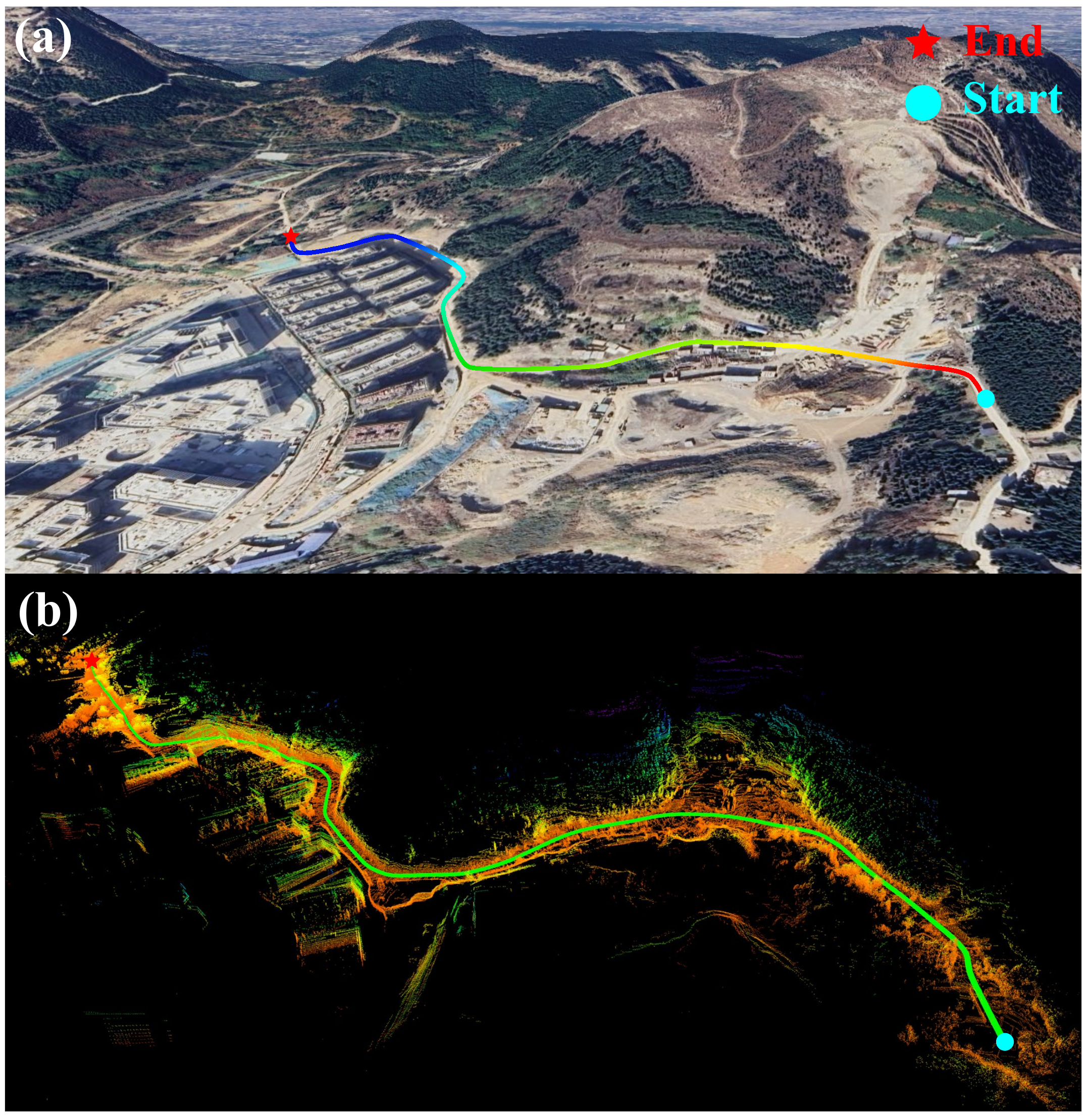}}
\caption{Mapping results in \textit{Offroad2} dataset. (a) is the satellite map from Google Earth. (b) shows the point cloud and trajectory.}
\label{fig: off12}
\end{figure}

\subsection{Mapping Results}

In this section, we focus on the mapping quality of different methods, which qualitatively reflects the localization accuracy on the global scale.
Fig. \ref{fig: xls_mapping}(a)-(b) shows the satellite map and mapping result on \textit{Xinglong} dataset. 
Fig. \ref{fig: xls_mapping}(c) is a snapshot of a curved road with slopes. The green line shows the navigation trajectory.
The point cloud map depict the outline of the real environment, which showcases the high localization accuracy and mapping quality using our method. 

To demonstrate the performance of mapping in the uneven terrain scenes, we compare the mapping performance of \textbf{HDL-SLAM}, \textbf{LeGO-LOAM}, and \textbf{ROLO} using the \textit{Qianfo} dataset. 
The results are shown in Fig. \ref{fig: qfs_mapping}(b)-(d). At a global scale, the point cloud map generated by our method exhibits high consistency with the scene contours in the satellite map.
In contrast, in the top figures of Fig. \ref{fig: qfs_mapping}(b)-(c), the results from \textbf{HDL-SLAM} and \textbf{LeGO-LOAM} suffer from noticeable shape distortion, especially in the contours of buildings.
In the side views of Fig. \ref{fig: qfs_mapping}(d), \textbf{ROLO} demonstrates a relatively flat mapping plane, given the level ground of our campus. 
However, the side views of other methods display significant oblique, and \textbf{HDL-LOAM} even show an overlap in Fig. \ref{fig: qfs_mapping}(b).
These phenomena reflect the significant vertical drifts in SLAM tasks. In contrast, \textbf{ROLO} has better performance in alleviating the vertical drift problem. 
Fig. \ref{fig: two_slope}(a)-(d) show two slope scenes in \textit{Qianfo} datasets. Our method can exhibit the correct and smooth trajectory and reconstruct original scenes with slopes, as shown in Fig. \ref{fig: two_slope}(c)-(d).

\begin{figure*}[t]
\centering
\includegraphics[scale=0.345]{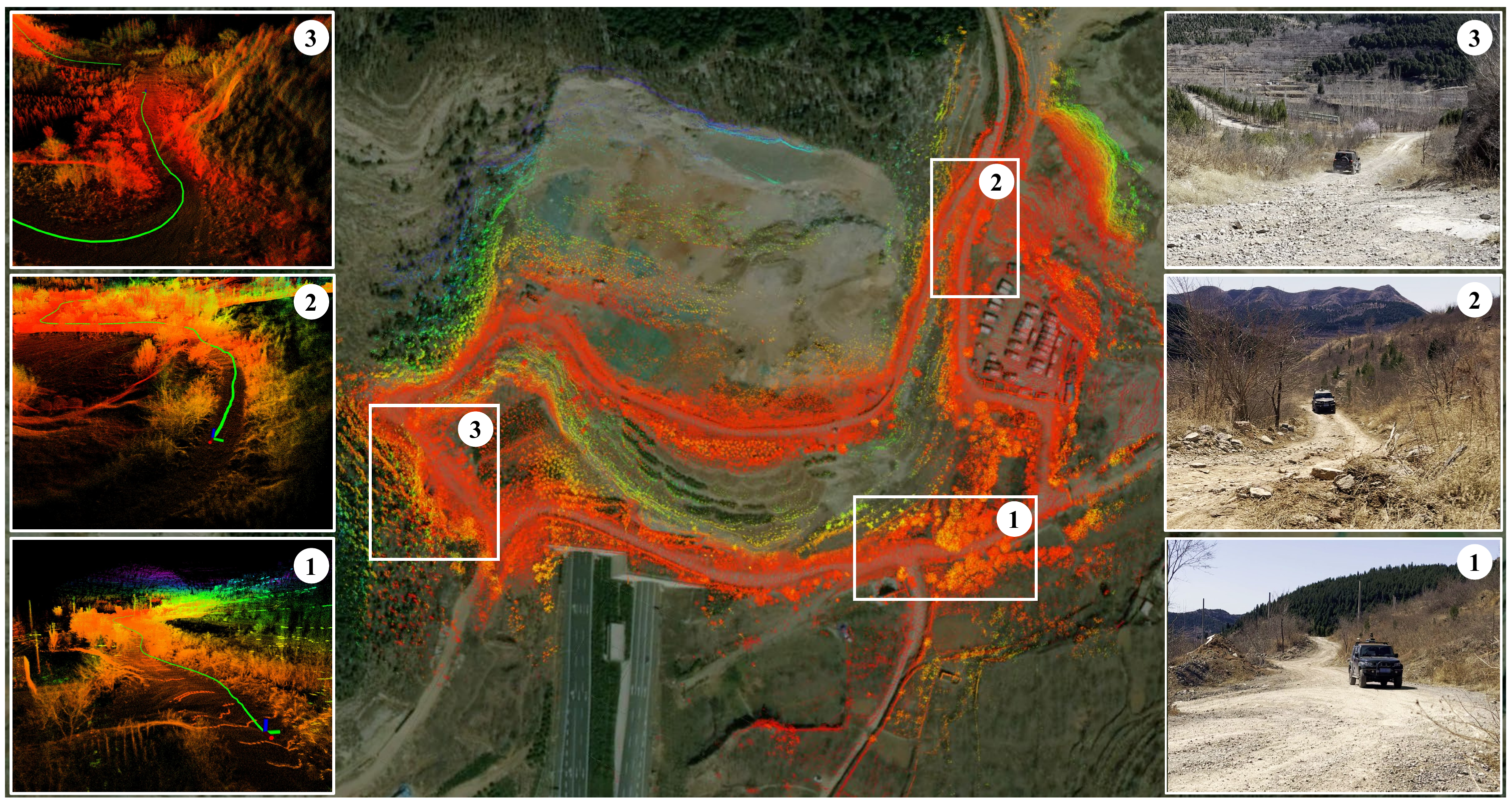}
\caption{Point cloud map aligned with satellite map of \textit{Offroad3} dataset. Three snapshots are captured in different driving scenes, which are sequenced by the top number.}
\label{fig: off3_mapping}
\end{figure*}

Regarding the off-road cases, our method retains promising performance in mapping quality. 
Fig. \ref{fig: off12}(a) shows a satellite map in \textit{Offroad2}, which exhibits a rugged downhill. The navigation trajectory and mapping results using \textbf{ROLO} are shown in Fig. \ref{fig: off12}(b). The gradient color of the trajectory indicates the elevation change, which matches the changes in the real world.  
It can be observed that the surrounding scenes are reconstructed accurately, such as forests and buildings.
To be more specific,  Fig. \ref{fig: off3_mapping} displays the point cloud map aligned with a satellite map and six specific snapshots of scenes and point cloud maps.
We can observe that the point cloud map coincides with the terrain contours in the satellite map. 
The right three snapshots show the real-world navigation scenes. The trajectories and mapping results are located on the left, which both exhibit high mapping quality as well as localization accuracy. 
Overall, \textbf{ROLO} has high efficiency and efficacy in robust and real-time localization and mapping, particularly in reducing vertical drift. These advantages enable \textbf{ROLO} applicable in challenging environments, such as autonomous off-road terrain navigation. 
\begin{figure*}[p]
	\setlength{\abovecaptionskip}{-2pt}
	\setlength{\belowcaptionskip}{-10pt}
	\centering
        \includegraphics[scale=0.35]{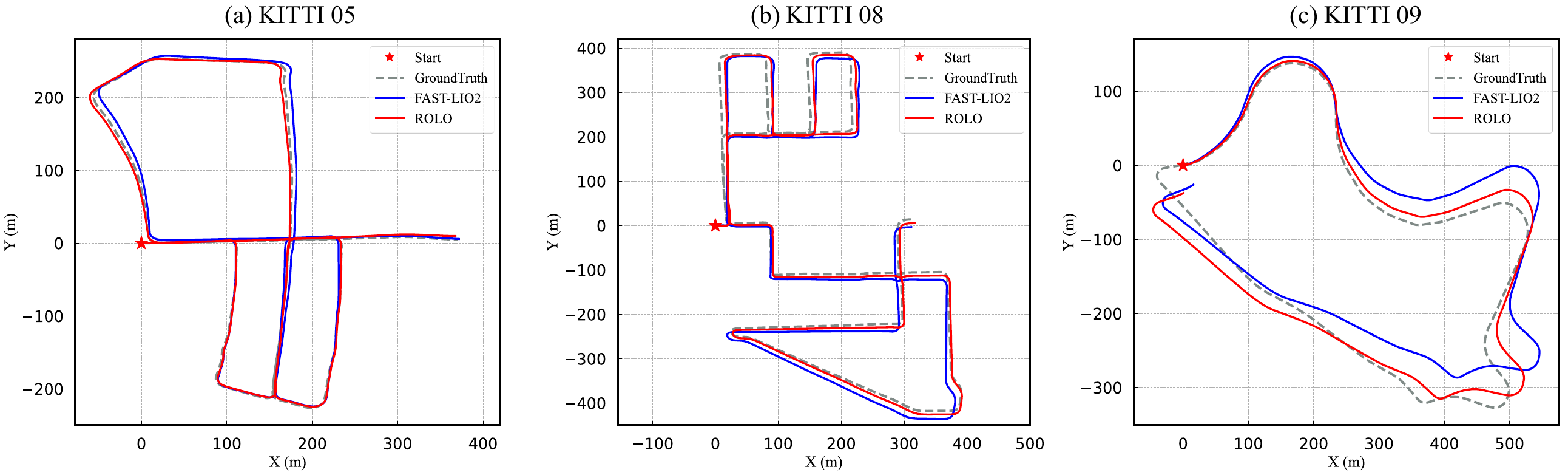}
        \vspace{0.1cm}
	\caption{Trajectory results of our method and \textbf{FAST-LIO2} for the \textit{KITTI sequence 05, 08, 09.}}
	\label{fig: kitti_trajectory_lio}
\end{figure*}
\begin{figure*}[p]
	\setlength{\abovecaptionskip}{-2pt}
	\setlength{\belowcaptionskip}{-10pt}
	\centering
        \includegraphics[scale=0.29]{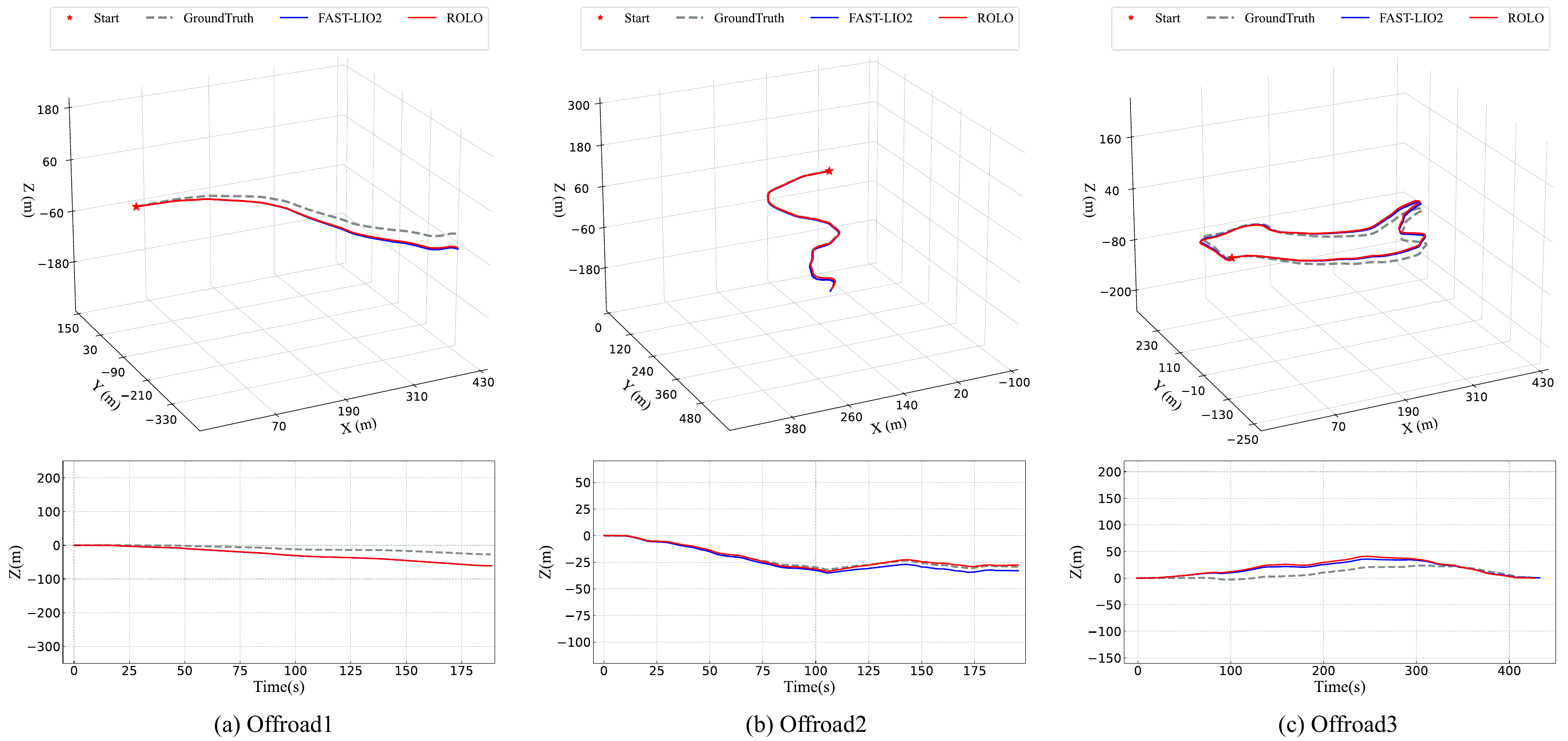}
        \vspace{0.1cm}
	\caption{The trajectory estimation of our method and \textbf{FAST-LIO2} for the datasets under real off-road scenes (\textit{Offroad1}, \textit{Offroad2}, \textit{Offroad3}).}
	\label{fig: off_trajectory_lio}
\end{figure*}
\begin{figure*}[p]
	\setlength{\abovecaptionskip}{-2pt}
	\setlength{\belowcaptionskip}{-10pt}
	\centering
	\includegraphics[scale=0.34]{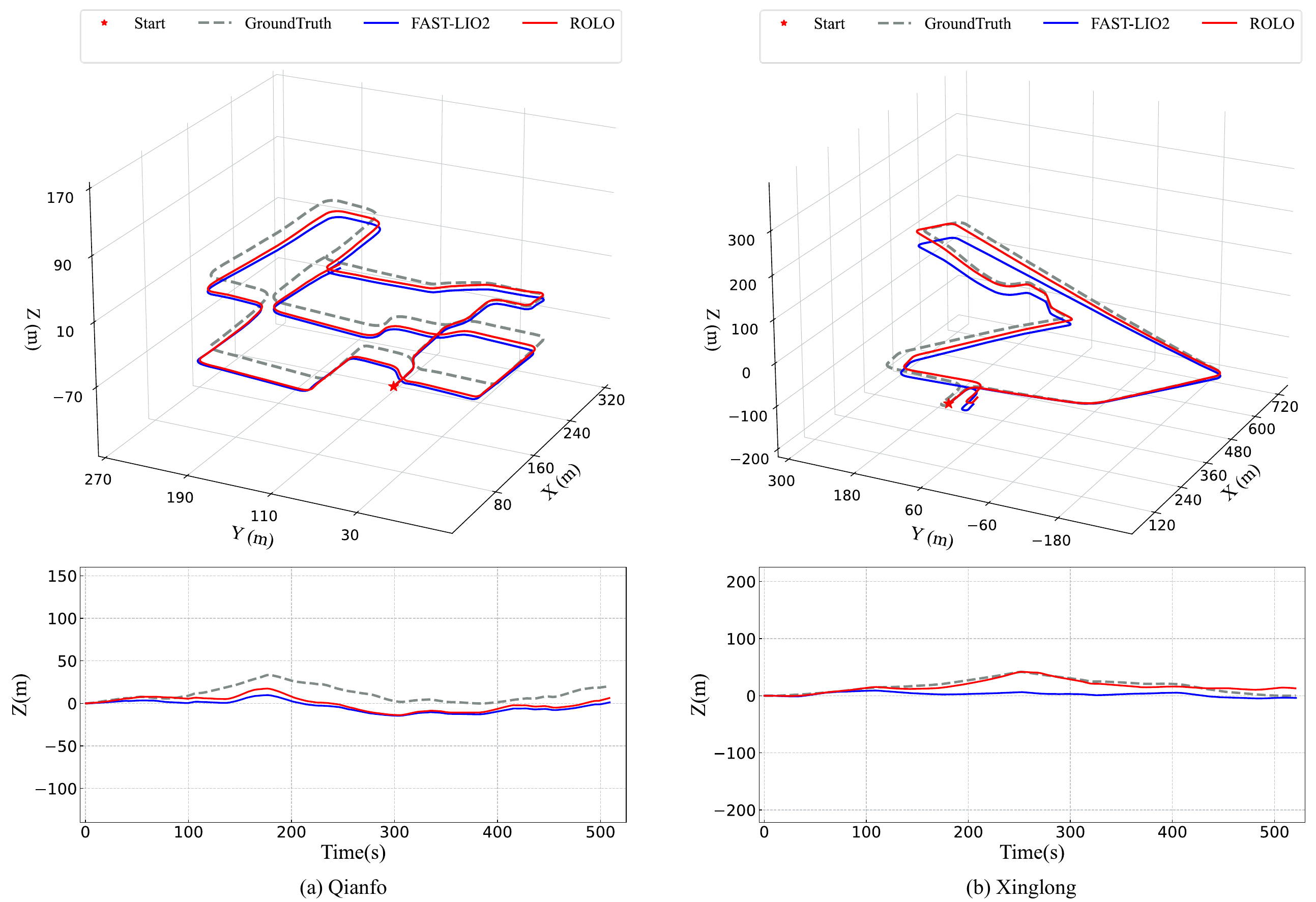}
        \vspace{0.1cm}
	\caption{The trajectory estimation of our method and \textbf{FAST-LIO2} for the SDU campus datasets (\textit{Qianfo}, \textit{Xinglong}).}
	\label{fig: sdu_trajectory_lio}
\end{figure*}
\subsection{Comparison with IMU-fused Method}
To further evaluate the localization performance of the proposed method, we compare it with \textbf{FAST-LIO2} in this section. Here, \textbf{FAST-LIO2} \cite{xu2022fastlio2} is a state-of-the-art LiDAR-inertial SLAM solution, which applies tightly coupled iterated Kalman filter and an \textit{ikd-tree} data structure to accomplish fast and efficient SLAM process. It is worth noting that \textbf{FAST-LIO2} compensates vehicle rotation through the highly frequent orientation observation from IMU, thus it has more excellent localization performance than other single LiDAR methods. 

We compare the localization accuracy, robustness and mapping results in the \textit{KITTI} datasets and the collected datasets under real off-road scene and SDU campus scene, respectively.
\begin{table*}[t]
\centering
\renewcommand\arraystretch{1.3}
\tabcolsep=0.14cm
\caption{RMSE (translation (m)/rotation (deg)) of \textbf{ROLO} and \textbf{FAST-LIO2} in evaluated datasets}
\label{tab: RMSE_lio}
\begin{tabular}{lccccccccc}
\hline
            & KITTI seq.00 & KITTI seq.05 & KITTI seq.08 & KITTI seq.09 & \begin{tabular}[c]{@{}c@{}}SDU camp\\ (Qianfo)\end{tabular} & \begin{tabular}[c]{@{}c@{}}SDU camp\\ (Xinglong)\end{tabular} & Offroad1 & Offroad2 & Offroad3 \\ \hline
ROLO      & \textbf{0.2}/\textbf{2.369} & \textbf{0.085}/\textbf{2.729} & \textbf{0.291}/3.045 & \textbf{0.852}/3.705 & \textbf{0.446}/\textbf{0.149}  & \textbf{0.348}/\textbf{0.046} & \textbf{0.524}/0.552 & \textbf{0.056}/0.043 & \textbf{0.269}/0.21  \\
FAST-LIO2 & fail/fail   & 0.134/2.863 & 0.374/\textbf{2.418} & 1.099/\textbf{3.351} & 0.544/0.158 & 0.525/0.076 & 0.988/\textbf{0.179} & 0.115/\textbf{0.025} & 0.309/\textbf{0.194} \\ \hline
\end{tabular}
\end{table*}
\begin{table*}[t]
\centering
\renewcommand\arraystretch{1.3}
\tabcolsep=0.13cm
\caption{Mean value of absolute error for inter-frame towards evaluated datasets (data format: $ \mathcal{\mathbf{E}}^{\mathbf{t}}/ \mathcal{\mathbf{E}}^{\mathbf{R}}$, unit: m/deg)}
\label{tab: absolute_error_lio}
\begin{tabular}{lccccccccc}
\hline
            & KITTI seq.00 & KITTI seq.05 & KITTI seq.08 & KITTI seq.09 & \begin{tabular}[c]{@{}c@{}}SDU camp\\ (Qianfo)\end{tabular} & \begin{tabular}[c]{@{}c@{}}SDU camp\\ (Xinglong)\end{tabular} & Offroad1 & Offroad2 & Offroad3 \\ \hline
ROLO      & \textbf{0.029}/\textbf{1.183} & \textbf{0.022}/\textbf{0.202}                      & \textbf{0.037}/\textbf{0.482}                      & 0.087/0.529                      & 0.057/0.135                                                                     & \textbf{0.130}/0.073                                                                       & 0.042/0.09                   & 0.031/0.114                  & 0.086/0.419                  \\
FAST-LIO2 & fail/fail                       & 0.028/0.217                      & 0.039/1.04                       & \textbf{0.080}/\textbf{0.290}                      & \textbf{0.049}/\textbf{0.069}                                                                     & 0.133/\textbf{0.053}                                                                       & \textbf{0.035}/\textbf{0.064}                  & \textbf{0.014}/\textbf{0.06}                   & \textbf{0.025}/\textbf{0.345}                  \\ \hline

\end{tabular}
\end{table*}
In \textit{KITTI} datasets, we compare our method and \textbf{FAST-LIO2} in \textit{KITTI seq. 05, 08, 09}, where \textbf{FAST-LIO2} works properly. 
The estimated trajectories and GT are depicted in Fig. \ref{fig: kitti_trajectory_lio}. \textbf{ROLO} exhibits closer distance than \textbf{FAST-LIO2} from GT in Fig. \ref{fig: kitti_trajectory_lio}(a)-(b), and \textbf{ROLO} particularly exhibits smaller cumulative error in Fig. \ref{fig: kitti_trajectory_lio}(c). In the \textit{Off-road} and \textit{SDU} datasets, the trajectory estimation results are shown as Fig. \ref{fig: off_trajectory_lio} and Fig. \ref{fig: sdu_trajectory_lio}. Intuitively, Fig. \ref{fig: off_trajectory_lio}(b) and Fig. \ref{fig: sdu_trajectory_lio}(a)-(b) showcase that comparing with \textbf{FAST-LIO2}, \textbf{ROLO} trajectory has less gap with GT. This phenomenon is pronounced in the vertical direction. \textbf{ROLO} is almost overlapped with GT. In Fig. \ref{fig: off_trajectory_lio}(a) and Fig. \ref{fig: off_trajectory_lio}(c), the trajectories of the two methods have high coincidence. Overall, the trajectory results indicate these methods have similar localization performance in uneven scenes, while \textbf{ROLO} has the better effect in \textit{KITTI seq. 05, 08, 09}. 

Tab. \ref{tab: RMSE_lio} records the RMSE of \textbf{ROLO} and \textbf{FAST-LIO2} for translation and rotation in datasets. \textbf{ROLO} has less translation RMSE, while \textbf{FAST-LIO2} has less rotation RMSE in both \textit{KITTI} and \textit{Off-road} datasets. Specifically, \textbf{ROLO} exhibits minimal errors than \textbf{FAST-LIO2} in \textit{SDU} datasets. It indicates that \textbf{ROLO} has higher accuracy than \textbf{FAST-LIO2} in position estimation. \textbf{FAST-LIO2} with the aid of direct attitude measurement of IMU obtains better attitude estimation performance.
\begin{figure*}[t]
\centering
\includegraphics[scale=0.43]{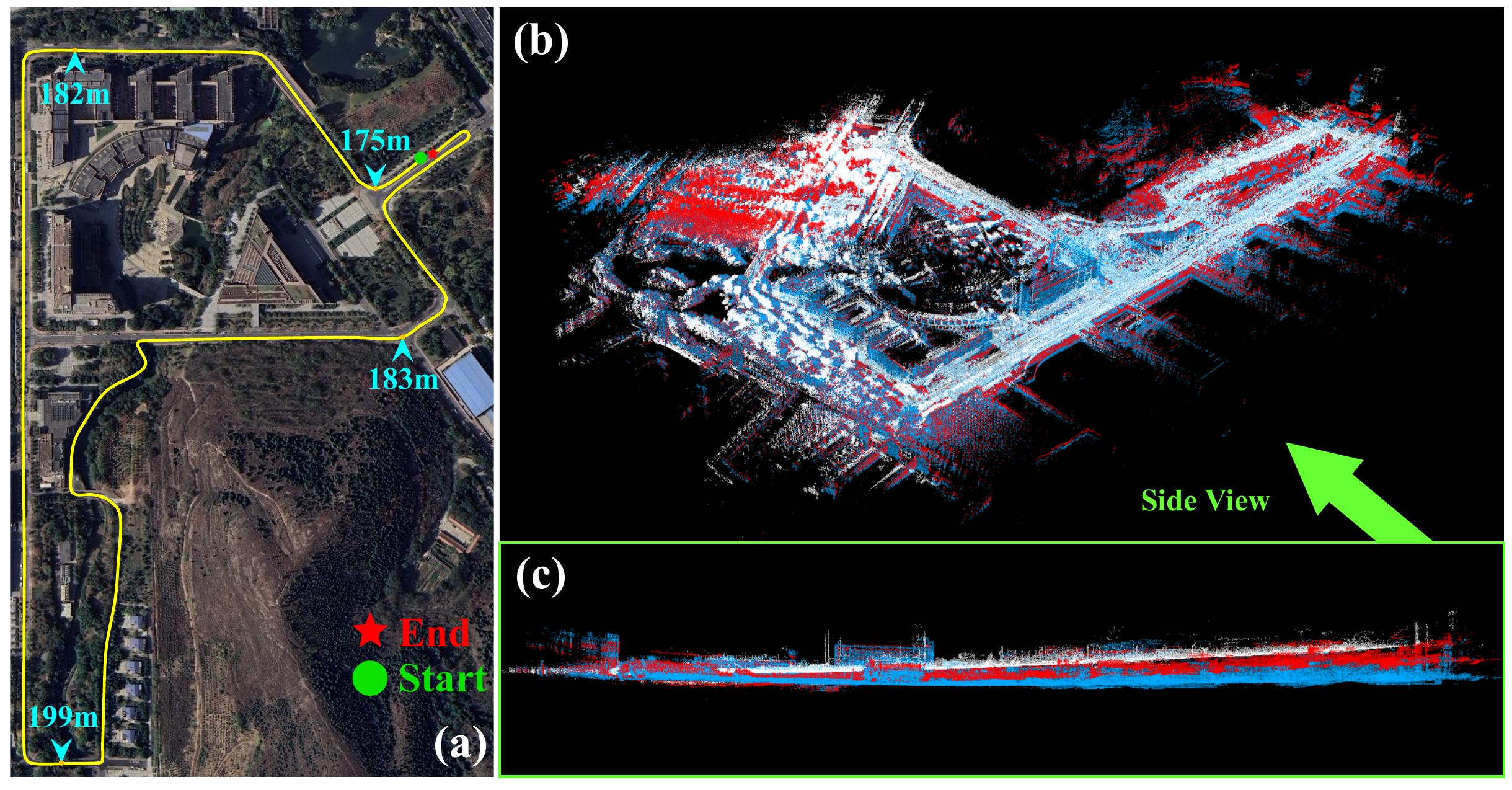}
\caption{Mapping results on \textit{Xinglong} dataset. (a) shows the satellite map from Google Earth, where light blue landmarks show absolute altitudes measured by a handheld altimeter. (b) shows the point cloud map, where the white, red and blue point clouds are generated by GT, \textbf{ROLO} and \textbf{FAST-LIO2}, respectively. (c) shows a side view of point cloud maps.}
\label{fig: comp_lio}
\end{figure*}
To showcase the localization robustness of two methods, we calculate the absolute error magnitudes, denoted as $\left \| \mathcal{\mathbf{E}}^{\mathbf{t} } \right \|$ and $\left \| \mathcal{\mathbf{E}}^{\mathbf{R} } \right \|$, in accordance with Eq. \ref{eq: absolute error_t} and Eq. \ref{eq: absolute error_r}. The results are shown in Tab. \ref{tab: absolute_error_lio}. \textbf{ROLO} has smaller mean values of $\left \| \mathcal{\mathbf{E}}^{\mathbf{t}} \right \|$ and $\left \| \mathcal{\mathbf{E}}^{\mathbf{R}} \right \|$ than \textbf{FAST-LIO2} in \textit{KITTI} datasets. In uneven terrains, \textbf{FAST-LIO2} shows better performance of localization robustness in the rotation domain. These results indicate that our method exhibits similar localization robustness to \textbf{FAST-LIO2} on the whole, even though the two methods demonstrate their strength in different scenarios. Fig. \ref{fig: comp_lio} shows the mapping results of two methods in \textit{Xinglong} dataset. It demonstrates that the point cloud of \textbf{ROLO} exhibits greater overlap than \textbf{FAST-LIO2}. In Fig. \ref{fig: comp_lio}(c), the ground plane of our method is closer to GT, and its slope is able to reflect the variation of altitude referred to the landmarks of Fig. \ref{fig: comp_lio}(a). The results show that \textbf{ROLO} is able to achieve a similar and even better localization performance than \textbf{FAST-LIO2} without the assistance of IMU, particularly in alleviating vertical drift.

In addition, the comparison results of computation efficiency between LiDAR-only methods and \textbf{FAST-LIO2} are shown in Fig. \ref{fig: time_box}. The results intuitively exhibit that time boxes of \textbf{FAST-LIO2} are located at the lowest location in datasets, which indicates the high performance of computation efficiency. It mainly benefits from fast data integration and high-frequency IMU measurement. Although \textbf{FAST-LIO2} has excellent computation efficiency, it cannot work properly in the absence of IMU, such as scenarios in \textit{KITTI seq. 00}. Yet, \textbf{ROLO} executes tasks only relies on the single LiDAR and has similar localization performance with \textbf{FAST-LIO2}.

\section{Discussions}
\label{result analysis}
The experimental results demonstrate that our method has convincing performance in attitude estimation and vertical drift mitigation. 
ROLO-SLAM leverages independent rotation and translation estimation at the front-end, where the rotation estimation applies the spherical alignment to achieve precise attitude alignment while the translation estimation considers the geometric and continuous-time constraints. 
The front-end offers reasonable pose guesses for the back-end pose optimization. The final translation and rotation errors are less than 0.6 meters and 0.6 degrees in uneven terrain, respectively. 
In addition, our method still maintains real-time performance though the front-end function is divided and rearranged. 
CT-ICP holds similar performance with ROLO-SLAM in accuracy and robustness.
However, its processing time for each scan is over 150 ms, while ROLO-SLAM is less than 120 ms. 
The reason is that, without incorporating global optimization, CT-ICP solely relies on scan-to-map residuals to constrain the transformation for geometric alignment. This process involves frequent point cloud trimming operations and matching, which significantly prolongs the processing time. 
From the above experiments, it can be concluded that our method has promising performance in off-road outdoor environments. 
Our method demonstrates a significant enhancement in localization accuracy, achieving an improvement of $45.8\%$ over LeGO-LOAM. Our method, without IMU assistance, achieves a similar localization performance by comparing with FAST-LIO2, while FAST-LIO2 has better real-time processing performance.
Moreover, the rough terrains are reconstructed more accurately with our method relying on precise localization. 

We conduct numerous experiments to evaluate the localization and mapping performance of ROLO-SLAM. Most results prove that our method has better performance in alleviating vertical drift compared with the state of the arts. However, we notice that the vertical drift is still unavoidable in some scenarios, such as vehicles driving on congested roads or competing in off-road rally environments. The reason is that the vehicle might have unpredictable motion behaviors or suffer emergency braking in these scenes, which leads to a worse solution in the forward location prediction. The accuracy of ROLO-SLAM is now reliant on forward location prediction. To break this constraint, we will reduce the dependence on forward location prediction and study the constraint-free decoupling between rotation and translation in future work. Extracting prior environmental information, like slope and time-sequenced information, is expected to be integrated into the future framework.

\section{Conclusions and Future Work}
\label{conclusion}
In this article, we propose the ROLO-SLAM to obtain precise pose estimation and environmental map in uneven terrain. To alleviate the vertical drift during ground vehicle driving, we independently estimate rotation and translation based on the coarse translation from the forward location prediction. In the back-end, we leverage the scan-to-submap and factor graph to promote the accuracy of the final pose estimation. The experimental results demonstrate that our method performs much better in the comparison with the state-of-the-art methods. Furthermore, the mapping evaluation in various scenes is conducted to validate the mapping quality of ROLO-SLAM. The mapping results show that our method is able to produce highly accurate point cloud maps in both urban and off-road scenes. Comparing the state-of-the-art methods, point cloud maps output by our method have the highest similarity with the real scenes. This also reflects that our method can output precise pose estimation.

In future work, we will study decoupling between the rotation and translation without the center-aligned condition, which reduces the dependence on the forward location prediction. Furthermore, we will focus on attitude mutation aroused by the dynamic and momentary ground fluctuations. Prior information like future pose prediction extracted from limited observation might be fused into the framework for improving the accuracy of attitude estimation.

\bibliographystyle{IEEEtran}
\bibliography{reference}

\begin{thebibliography}{10}
\providecommand{\url}[1]{#1}
\csname url@samestyle\endcsname
\providecommand{\newblock}{\relax}
\providecommand{\bibinfo}[2]{#2}
\providecommand{\BIBentrySTDinterwordspacing}{\spaceskip=0pt\relax}
\providecommand{\BIBentryALTinterwordstretchfactor}{4}
\providecommand{\BIBentryALTinterwordspacing}{\spaceskip=\fontdimen2\font plus
\BIBentryALTinterwordstretchfactor\fontdimen3\font minus \fontdimen4\font\relax}
\providecommand{\BIBforeignlanguage}[2]{{%
\expandafter\ifx\csname l@#1\endcsname\relax
\typeout{** WARNING: IEEEtran.bst: No hyphenation pattern has been}%
\typeout{** loaded for the language `#1'. Using the pattern for}%
\typeout{** the default language instead.}%
\else
\language=\csname l@#1\endcsname
\fi
#2}}
\providecommand{\BIBdecl}{\relax}
\BIBdecl

\bibitem{latif2014robust}
Y.~Latif, C.~Cadena, and J.~Neira, ``Robust graph slam back-ends: A comparative analysis,'' in \emph{2014 IEEE/RSJ International Conference on Intelligent Robots and Systems}.\hskip 1em plus 0.5em minus 0.4em\relax IEEE, 2014, pp. 2683--2690, \url{https://doi.org/10.1109/IROS.2014.6942929}.

\bibitem{lego_loam}
T.~Shan and B.~Englot, ``Lego-loam: Lightweight and ground-optimized lidar odometry and mapping on variable terrain,'' in \emph{2018 IEEE/RSJ International Conference on Intelligent Robots and Systems (IROS)}.\hskip 1em plus 0.5em minus 0.4em\relax IEEE, 2018, pp. 4758--4765, \url{https://doi.org/10.1109/IROS.2018.8594299}.

\bibitem{galeote2023gnd}
A.~Galeote-Luque, J.-R. Ruiz-Sarmiento, and J.~Gonzalez-Jimenez, ``Gnd-lo: Ground decoupled 3d lidar odometry based on planar patches,'' \emph{IEEE Robotics and Automation Letters}, 2023, \url{https://doi.org/10.1109/LRA.2023.3313057}.

\bibitem{chen2024ig}
Z.~Chen, Y.~Xu, S.~Yuan, and L.~Xie, ``ig-lio: An incremental gicp-based tightly-coupled lidar-inertial odometry,'' \emph{IEEE Robotics and Automation Letters}, 2024, \url{https://doi.org/10.1109/LRA.2024.3349915}.

\bibitem{jian2022putn}
Z.~Jian, Z.~Lu, X.~Zhou, B.~Lan, A.~Xiao, X.~Wang, and B.~Liang, ``Putn: A plane-fitting based uneven terrain navigation framework,'' in \emph{2022 IEEE/RSJ International Conference on Intelligent Robots and Systems (IROS)}.\hskip 1em plus 0.5em minus 0.4em\relax IEEE, 2022, pp. 7160--7166, \url{https://doi.org/10.1109/IROS47612.2022.9981038}.

\bibitem{ebadi2020lamp}
K.~Ebadi, Y.~Chang, M.~Palieri, A.~Stephens, A.~Hatteland, E.~Heiden, A.~Thakur, N.~Funabiki, B.~Morrell, S.~Wood \emph{et~al.}, ``Lamp: Large-scale autonomous mapping and positioning for exploration of perceptually-degraded subterranean environments,'' in \emph{2020 IEEE International Conference on Robotics and Automation (ICRA)}.\hskip 1em plus 0.5em minus 0.4em\relax IEEE, 2020, pp. 80--86, \url{https://doi.org/10.1109/ICRA40945.2020.9197082}.

\bibitem{xue2023traversability}
H.~Xue, H.~Fu, L.~Xiao, Y.~Fan, D.~Zhao, and B.~Dai, ``Traversability analysis for autonomous driving in complex environment: A lidar-based terrain modeling approach,'' \emph{Journal of Field Robotics}, vol.~40, no.~7, pp. 1779--1803, 2023, \url{https://doi.org/10.1002/rob.22209}.

\bibitem{2014loam}
J.~Zhang and S.~Singh, ``Loam: Lidar odometry and mapping in real-time.'' in \emph{Robotics: Science and systems}, vol.~2, no.~9.\hskip 1em plus 0.5em minus 0.4em\relax Berkeley, CA, 2014, pp. 1--9, \url{https://doi.org/10.15607/RSS.2014.X.007}.

\bibitem{wang2021floam}
H.~Wang, C.~Wang, C.-L. Chen, and L.~Xie, ``F-loam: Fast lidar odometry and mapping,'' in \emph{2021 IEEE/RSJ International Conference on Intelligent Robots and Systems (IROS)}.\hskip 1em plus 0.5em minus 0.4em\relax IEEE, 2021, pp. 4390--4396, \url{https://doi.org/10.1109/IROS51168.2021.9636655}.

\bibitem{lin2020livox_loam}
J.~Lin and F.~Zhang, ``Loam livox: A fast, robust, high-precision lidar odometry and mapping package for lidars of small fov,'' in \emph{2020 IEEE International Conference on Robotics and Automation (ICRA)}.\hskip 1em plus 0.5em minus 0.4em\relax IEEE, 2020, pp. 3126--3131, \url{https://doi.org/10.1109/ICRA40945.2020.9197440}.

\bibitem{chen2022direct}
K.~Chen, B.~T. Lopez, A.-a. Agha-mohammadi, and A.~Mehta, ``Direct lidar odometry: Fast localization with dense point clouds,'' \emph{IEEE Robotics and Automation Letters}, vol.~7, no.~2, pp. 2000--2007, 2022, \url{https://doi.org/10.1109/LRA.2022.3142739}.

\bibitem{pomerleau2013comparing}
F.~Pomerleau, F.~Colas, R.~Siegwart, and S.~Magnenat, ``Comparing icp variants on real-world data sets: Open-source library and experimental protocol,'' \emph{Autonomous robots}, vol.~34, pp. 133--148, 2013, \url{https://doi.org/10.1109/IM.2001.924446}.

\bibitem{li2022gicp_loam}
X.~Li, A.~Zhang, H.~Liu, C.~Fan, C.~Liu, and R.~Zheng, ``Gicp-loam: Lidar odometry and mapping with voxelized generalized iterative closest point,'' in \emph{2022 China Automation Congress (CAC)}.\hskip 1em plus 0.5em minus 0.4em\relax IEEE, 2022, pp. 2103--2108, \url{https://doi.org/10.1109/CAC57257.2022.10055138}.

\bibitem{chen2021ndtloam}
S.~Chen, H.~Ma, C.~Jiang, B.~Zhou, W.~Xue, Z.~Xiao, and Q.~Li, ``Ndt-loam: A real-time lidar odometry and mapping with weighted ndt and lfa,'' \emph{IEEE Sensors Journal}, vol.~22, no.~4, pp. 3660--3671, 2021, \url{https://doi.org/10.1109/JSEN.2021.3135055}.

\bibitem{dellenbach2022cticp}
P.~Dellenbach, J.-E. Deschaud, B.~Jacquet, and F.~Goulette, ``Ct-icp: Real-time elastic lidar odometry with loop closure,'' in \emph{2022 International Conference on Robotics and Automation (ICRA)}.\hskip 1em plus 0.5em minus 0.4em\relax IEEE, 2022, pp. 5580--5586, \url{https://doi.org/10.1109/ICRA46639.2022.9811849}.

\bibitem{2023fastfeature}
S.~Choi, H.-W. Chae, Y.~Jeung, S.~Kim, K.~Cho, and T.-w. Kim, ``Fast and versatile feature-based lidar odometry via efficient local quadratic surface approximation,'' \emph{IEEE Robotics and Automation Letters}, vol.~8, no.~2, pp. 640--647, 2023, \url{https://doi.org/10.1109/LRA.2022.3227875}.

\bibitem{2023localinformation}
H.~Guo, J.~Zhu, and Y.~Chen, ``E-loam: Lidar odometry and mapping with expanded local structural information,'' \emph{IEEE Transactions on Intelligent Vehicles}, vol.~8, no.~2, pp. 1911--1921, 2023, \url{https://doi.org/10.1109/TIV.2022.3151665}.

\bibitem{2022linefeature}
S.~Guo, Z.~Rong, S.~Wang, and Y.~Wu, ``A lidar slam with pca-based feature extraction and two-stage matching,'' \emph{IEEE Transactions on Instrumentation and Measurement}, vol.~71, pp. 1--11, 2022, \url{https://doi.org/10.1109/TIM.2022.3156982}.

\bibitem{wang2022fevo}
Z.~Wang, L.~Yang, F.~Gao, and L.~Wang, ``Fevo-loam: Feature extraction and vertical optimized lidar odometry and mapping,'' \emph{IEEE Robotics and Automation Letters}, vol.~7, no.~4, pp. 12\,086--12\,093, 2022, \url{https://doi.org/10.1109/LRA.2022.3201689}.

\bibitem{chen2022low}
X.~Chen, Y.~Wang, C.~Wang, R.~Song, and Y.~Li, ``Low-drift lidar-only odometry and mapping for ugvs in environments with non-level roads,'' in \emph{2022 IEEE/RSJ International Conference on Intelligent Robots and Systems (IROS)}.\hskip 1em plus 0.5em minus 0.4em\relax IEEE, 2022, pp. 13\,174--13\,180, \url{https://doi.org/10.1109/IROS47612.2022.9982264}.

\bibitem{yang2020teaser}
H.~Yang, J.~Shi, and L.~Carlone, ``Teaser: Fast and certifiable point cloud registration,'' \emph{IEEE Transactions on Robotics}, vol.~37, no.~2, pp. 314--333, 2020, \url{https://doi.org/10.1109/TRO.2020.3033695}.

\bibitem{chen2022eil}
W.~Chen, Y.~Wang, H.~Chen, and Y.~Liu, ``Eil-slam: Depth-enhanced edge-based infrared-lidar slam,'' \emph{Journal of Field Robotics}, vol.~39, no.~2, pp. 117--130, 2022, \url{https://doi.org/10.1002/rob.22040}.

\bibitem{wang2022dlio}
Z.~Wang, L.~Zhang, Y.~Shen, and Y.~Zhou, ``D-liom: Tightly-coupled direct lidar-inertial odometry and mapping,'' \emph{IEEE Transactions on Multimedia}, 2022, \url{https://doi.org/10.1109/TMM.2022.3168423}.

\bibitem{shan2020lio}
T.~Shan, B.~Englot, D.~Meyers, W.~Wang, C.~Ratti, and D.~Rus, ``Lio-sam: Tightly-coupled lidar inertial odometry via smoothing and mapping,'' in \emph{2020 IEEE/RSJ international conference on intelligent robots and systems (IROS)}.\hskip 1em plus 0.5em minus 0.4em\relax IEEE, 2020, pp. 5135--5142, \url{https://doi.org/10.1109/IROS45743.2020.9341176}.

\bibitem{2023sdv}
Z.~Yuan, Q.~Wang, K.~Cheng, T.~Hao, and X.~Yang, ``Sdv-loam: Semi-direct visual-lidar odometry and mapping,'' \emph{IEEE Transactions on Pattern Analysis and Machine Intelligence}, pp. 1--18, 2023, \url{https://doi.org/10.1109/TPAMI.2023.3262817}.

\bibitem{pais20203dregnet}
G.~D. Pais, S.~Ramalingam, V.~M. Govindu, J.~C. Nascimento, R.~Chellappa, and P.~Miraldo, ``3dregnet: A deep neural network for 3d point registration,'' in \emph{Proceedings of the IEEE/CVF conference on computer vision and pattern recognition}, 2020, pp. 7193--7203, \url{https://doi.org/10.1109/CVPR42600.2020.00722}.

\bibitem{li2019net}
Q.~Li, S.~Chen, C.~Wang, X.~Li, C.~Wen, M.~Cheng, and J.~Li, ``Lo-net: Deep real-time lidar odometry,'' in \emph{Proceedings of the IEEE/CVF Conference on Computer Vision and Pattern Recognition}, 2019, pp. 8473--8482, \url{https://doi.org/10.1109/CVPR.2019.00867}.

\bibitem{chen2020overlapnet}
X.~Chen, T.~L\"abe, A.~Milioto, T.~R\"ohling, O.~Vysotska, A.~Haag, J.~Behley, and C.~Stachniss, ``{OverlapNet: Loop Closing for LiDAR-based SLAM},'' in \emph{Proceedings of Robotics: Science and Systems (RSS)}, 2020, \url{https://doi.org/10.15607/RSS.2020.XVI.009}.

\bibitem{chen2019suma++}
X.~Chen, A.~Milioto, E.~Palazzolo, P.~Giguere, J.~Behley, and C.~Stachniss, ``Suma++: Efficient lidar-based semantic slam,'' in \emph{2019 IEEE/RSJ International Conference on Intelligent Robots and Systems (IROS)}.\hskip 1em plus 0.5em minus 0.4em\relax IEEE, 2019, pp. 4530--4537, \url{https://doi.org/10.1109/IROS40897.2019.8967704}.

\bibitem{deng2023nerf}
J.~Deng, Q.~Wu, X.~Chen, S.~Xia, Z.~Sun, G.~Liu, W.~Yu, and L.~Pei, ``Nerf-loam: Neural implicit representation for large-scale incremental lidar odometry and mapping,'' in \emph{Proceedings of the IEEE/CVF International Conference on Computer Vision}, 2023, pp. 8218--8227, \url{https://doi.org/10.1109/ICCV51070.2023.00755}.

\bibitem{2023slamesh}
J.~Ruan, B.~Li, Y.~Wang, and Y.~Sun, ``Slamesh: Real-time lidar simultaneous localization and meshing,'' in \emph{2023 IEEE International Conference on Robotics and Automation (ICRA)}, 2023, pp. 3546--3552, \url{https://doi.org/10.1109/ICRA48891.2023.10161425}.

\bibitem{Geiger2012kitti}
A.~Geiger, P.~Lenz, and R.~Urtasun, ``Are we ready for autonomous driving? the kitti vision benchmark suite,'' in \emph{Conference on Computer Vision and Pattern Recognition (CVPR)}, 2012, \url{https://doi.org/10.1109/CVPR.2012.6248074}.

\bibitem{liu2023glio}
X.~Liu, W.~Wen, and L.-T. Hsu, ``Glio: Tightly-coupled gnss/lidar/imu integration for continuous and drift-free state estimation of intelligent vehicles in urban areas,'' \emph{IEEE Transactions on Intelligent Vehicles}, 2023, \url{https://doi.org/10.1109/TIV.2023.3323648}.

\bibitem{rehder2016kalibr}
J.~Rehder, J.~Nikolic, T.~Schneider, T.~Hinzmann, and R.~Siegwart, ``Extending kalibr: Calibrating the extrinsics of multiple imus and of individual axes,'' in \emph{2016 IEEE International Conference on Robotics and Automation (ICRA)}.\hskip 1em plus 0.5em minus 0.4em\relax IEEE, 2016, pp. 4304--4311, \url{https://doi.org/10.1109/ICRA.2016.7487628}.

\bibitem{kato2018autoware}
S.~Kato, S.~Tokunaga, Y.~Maruyama, S.~Maeda, M.~Hirabayashi, Y.~Kitsukawa, A.~Monrroy, T.~Ando, Y.~Fujii, and T.~Azumi, ``Autoware on board: Enabling autonomous vehicles with embedded systems,'' in \emph{2018 ACM/IEEE 9th International Conference on Cyber-Physical Systems (ICCPS)}.\hskip 1em plus 0.5em minus 0.4em\relax IEEE, 2018, pp. 287--296, \url{https://doi.org/10.1109/ICCPS.2018.00035}.

\bibitem{koide2019hdl_slam}
K.~Koide, J.~Miura, and E.~Menegatti, ``A portable three-dimensional lidar-based system for long-term and wide-area people behavior measurement,'' \emph{International Journal of Advanced Robotic Systems}, vol.~16, no.~2, p. 1729881419841532, 2019, \url{https://doi.org/10.1177/1729881419841532}.

\bibitem{xu2022fastlio2}
W.~Xu, Y.~Cai, D.~He, J.~Lin, and F.~Zhang, ``Fast-lio2: Fast direct lidar-inertial odometry,'' \emph{IEEE Transactions on Robotics}, vol.~38, no.~4, pp. 2053--2073, 2022, \url{https://doi.org/10.1109/TRO.2022.3141876}.

\end{thebibliography}

\end{document}